\documentclass[review]{elsarticle}
\graphicspath{ {./figures/} }

\usepackage{adjustbox}
\usepackage{hhline}
\usepackage{svg}
\usepackage{caption}
\usepackage{afterpage}
\usepackage{lipsum}
\usepackage{csquotes}

\usepackage{algpseudocode}
\usepackage{csquotes}
\usepackage{array}
\usepackage{textcomp}
\usepackage{stfloats}

\usepackage{subcaption}
\usepackage{siunitx}

\usepackage{pifont}

\usepackage{verbatim}
\usepackage{apalike}

\usepackage{ragged2e}
\usepackage[normalem]{ulem}
\usepackage{amsthm}
\usepackage[mathscr]{euscript}
\usepackage{algorithmicx}
\usepackage{algpseudocode}
\usepackage{stackengine}
\usepackage[linesnumbered,ruled,vlined]{algorithm2e}
\newcolumntype{P}[1]{>{\centering\arraybackslash}p{#1}}
\newcolumntype{M}[1]{>{\centering\arraybackslash}m{#1}}
\usepackage[fleqn]{amsmath}
\usepackage{bm}

\usepackage{mathtools}

\DeclareFontFamily{OT1}{pzc}{}
\DeclareFontShape{OT1}{pzc}{m}{it}{<-> s * [0.900] pzcmi7t}{}
\DeclareMathAlphabet{\mathpzc}{OT1}{pzc}{m}{it}
\usepackage{threeparttable}
\usepackage{booktabs}

\DeclareMathOperator*{\argmax}{argmax}

\usepackage{xcolor,colortbl}

\definecolor{gaincolor}{RGB}{230, 245, 255} 

\usepackage{footmisc}
\usepackage{scrextend}
\usepackage{graphicx}
\usepackage{amsfonts}
\usepackage{enumitem}
\usepackage[mathscr]{euscript}
\usepackage{mathtools}
\usepackage{amssymb}
\usepackage{array}
\usepackage{float}
\usepackage{threeparttable}
\usepackage{amsmath}
\usepackage{blindtext}
\usepackage{multirow}
\usepackage{color}
\usepackage{setspace}
\usepackage{hyperref}
\hypersetup{pdfauthor={Name}}
\usepackage{xcolor}

\usepackage[left=2.5cm,right=2.5cm,top=2cm,bottom=1.5cm]{geometry}

\newcommand{\xmark}{\ding{55}} 
\newcommand{\cmark}{\ding{51}} 

\journal{Information Fusion}

\biboptions{authoryear}

\begin{document}
\begin{frontmatter}

\begin{titlepage}
\begin{center}
\vspace*{1cm}

\textbf{ \large CLDTracker: A Comprehensive Language Description for Visual Tracking}

\vspace{1.5cm}

Mohamad Alansari$^{1}$ (100061914@ku.ac.ae), Sajid Javed$^{1}$ (sajid.javed@ku.ac.ae), Iyyakutti Iyappan Ganapathi$^{1}$ (iyyakutti.ganapathi@ku.ac.ae), Sara Alansari$^{1}$ (saraansari@live.com), and Muzammal Naseer$^{1}$ (muhammadmuzammal.naseer@ku.ac.ae) \\

\hspace{10pt}

\begin{flushleft}
\small  
$^{1}$ Department of Computer Science, Khalifa University, Abu Dhabi, United Arab Emirates. \\

\textbf{Corresponding Author:} \\
Mohamad Alansari \\
Department of Computer Science, Khalifa University, Abu Dhabi, United Arab Emirates. \\
Email: 100061914@ku.ac.ae

\end{flushleft}        
\end{center}
\end{titlepage}

\begin{abstract}
\noindent Visual Object Tracking (VOT) remains a fundamental yet challenging task in computer vision due to dynamic appearance changes, occlusions, and background clutter.
Traditional trackers, relying primarily on visual cues, often struggle in such complex scenarios. 
Recent advancements in Vision-Language Models (VLMs) have shown promise in semantic understanding for tasks like open-vocabulary detection and image captioning, suggesting their potential for VOT. 
However, the direct application of VLMs to VOT is hindered by critical limitations: the absence of a rich and comprehensive textual representation that semantically captures the target object’s nuances, limiting the effective use of language information; inefficient fusion mechanisms that fail to optimally integrate visual and textual features, preventing a holistic understanding of the target; and a lack of temporal modeling of the target’s evolving appearance in the language domain, leading to a disconnect between the initial description and the object’s subsequent visual changes. 
To bridge these gaps and unlock the full potential of VLMs for VOT, we propose CLDTracker, a novel \textbf{C}omprehensive \textbf{L}anguage \textbf{D}escription framework for robust visual \textbf{Track}ing. 
Our tracker introduces a dual-branch architecture consisting of a textual and a visual branch. 
In the textual branch, we construct a rich bag of textual descriptions derived by harnessing the powerful VLMs such as CLIP and GPT-4V, enriched with semantic and contextual cues to address the lack of rich textual representation. 
We further propose a \textbf{T}emporal \textbf{T}ext \textbf{F}eature \textbf{U}pdate \textbf{M}echanism (TTFUM) to adapt these descriptions across frames, capturing evolving target appearances and tackling the absence of temporal modeling.
In parallel, the visual branch extracts features using a Vision Transformer (ViT), and an attention-based cross-modal correlation head fuses both modalities for accurate target prediction, addressing the inefficient fusion mechanisms. 
Experiments on six standard VOT benchmarks demonstrate that CLDTracker achieves State-of-The-Art (SOTA) performance, validating the effectiveness of leveraging robust and temporally-adaptive vision-language representations for tracking.
Code and models are publicly available at: \url{https://github.com/HamadYA/CLDTracker}.
\end{abstract}

\end{frontmatter}

\newpage

\section{Introduction} \label{intro}
\noindent Visual Object Tracking (VOT) is a fundamental yet challenging problem in computer vision \cite{chen2022visual,vot_survey}.
The primary goal of VOT is to estimate the trajectory of a target object based on its initial position, provided as a bounding box or segmentation mask in the first frame \cite{stark, ostrack, mixformer, mixformer2, aiatrack, vot_survey}.
VOT has a wide range of applications, including video surveillance \cite{chen2022visual, reviewer_1}, human pose estimation \cite{pose}, robotic navigation and manipulation \cite{vasttrack}, and person re-identification \cite{dptuac}, among others.

Over the years, various VOT paradigms have been proposed, such as Discriminative Correlation Filters (DCF)-based trackers \cite{atom, kcf, dimp, prdimp}, Siamese-based trackers \cite{siamfc, siamrpn, siamrpn++, siamattn, siamban}, and Vision Transformer (ViT)-based trackers \cite{stark,ostrack,mixformer,videotrack,seqtrack}. 
These approaches have significantly advanced the field, aided by large-scale publicly available benchmark datasets \cite{lasot, lasotext, got10k, trackingnet}.
However, each category of these trackers comes with its strengths and limitations in the presence of occlusion, similar distractors, motion blur, fast motion, and out-of-plane rotation \cite{chen2022visual,vot_survey}.
The DCF-based trackers have been highly influential in the VOT due to their computational efficiency and competitive accuracy. However, these trackers struggle with boundary effects, limited robustness to appearance changes, overfitting to background, poor scale adaptation, model drift from online updates, and lack of semantic understanding—leading to reduced performance in dynamic or complex tracking scenarios \cite{atom, kcf, dimp, prdimp}.
The Siamese-based trackers are fast, end-to-end, and generalize well due to offline training on large datasets \cite{siamfc, siamrpn, siamrpn++, siamattn, siamban}.
They offer real-time performance without online updates, making them efficient and robust in standard scenarios. 
However, they suffer from template degradation over time, limited adaptability to appearance changes, and poor long-term tracking or re-detection.
ViT-based trackers excel at modeling global context and capturing long-range dependencies, leading to strong robustness and accuracy. However, they are computationally expensive and may require large-scale training data and resources for effective performance.

\begin{figure}[h!]
\centering    
\includegraphics[width=0.8\linewidth]{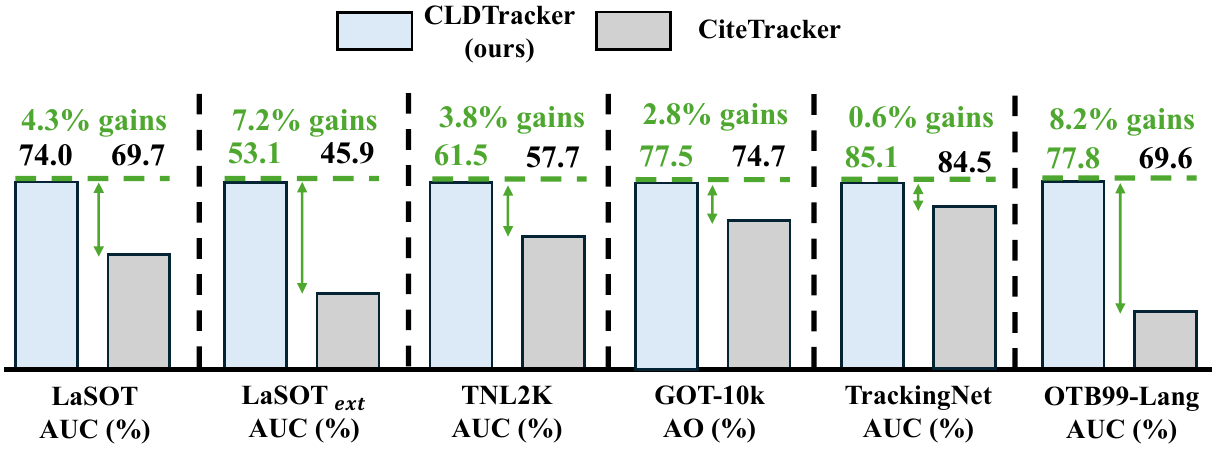}
\caption{VOT performance comparison in terms of Area Under the Curve (AUC) score between the CiteTracker \cite{citetracker} containing limited textual descriptions and our proposed CLDTracker containing comprehensive textual descriptions of the target object. CLDTracker shows significant performance improvements on six VOT benchmark datasets including LaSOT \cite{lasot}, LaSOT\textsubscript{ext} \cite{lasotext}, TNL2K \cite{tnl2k}, GOT-10k \cite{got10k}, TrackingNet \cite{trackingnet}, and OTB99-Lang \cite{lang_tracker}.}
\label{fig1}
\end{figure}


Multi-modal Vision-Language Models (VLMs) have revolutionized numerous computer vision applications, including open-vocabulary object detection, semantic segmentation, and visual grounding reasoning \cite{shi2024llmformer,awais2025foundation, cheng2024yolo}, among others.
In addition to the appearance-based features used in classical trackers, VLMs-based trackers leverage textual descriptions, contextual reasoning, and multi-modal embeddings to enhance VOT.
This enables better alignment and transferability between image and text features \cite{citetracker, jointnlt, vlt, divert, mmtrack, UVLTrack, context_aware}.
Though, VLM-based trackers leverage semantic understanding from text to improve robustness against appearance changes and enable category-agnostic tracking. 
They often rely on coarse language grounding, leading to limited spatial precision and high computational demands.
Thereofore, VLM-based trackers still lag behind SOTA visual trackers \cite{citetracker, jointnlt, vlt, divert, mmtrack, UVLTrack, context_aware, hiptrack, AQATrack, artrack} on VOT benchmarks such as LaSOT \cite{lasot, lasotext}, GOT-10K \cite{got10k}, and TrackingNet \cite{trackingnet}.
This performance gap is primarily attributed to two key issues.

\begin{figure*}[t]
  \centering
  \begin{subfigure}[b]{0.48\textwidth}
    \includegraphics[width=\textwidth]{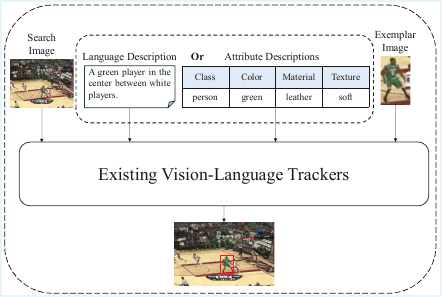}
    \caption{}
    \label{fig:1a}
  \end{subfigure}
  \begin{subfigure}[b]{0.48\textwidth}
    \includegraphics[width=\textwidth]{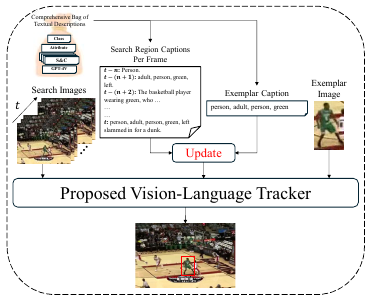}
    \caption{}
    \label{fig:1b}
  \end{subfigure}

\caption{Comparison of the visual-textual alignment process in the VLM-based trackers.
(a) show the traditional visual-textual alignment in the SOTA VLM-based trackers \cite{jointnlt,citetracker} and (b) shows our proposed CLDTracker containing a comprehensive bag of textual descriptions (\(B_t\)) which is then aligned with the visual representation and dynamic text generated from our Temporal Text Feature Update Mechanism (TTFUM).}

\label{fig2}

\end{figure*}

First, the publicly available benchmark datasets provide only a short phrase to describe the target object \cite{tnl2k, zhang2024webuot}.
As a result, the visual-textual alignment in the existing SOTA trackers is often limited, ambiguous, or inconsistent, reducing their ability to accurately focus on the target object (Fig. \ref{fig2}(a))\cite{vlt, divert}. 
Consequently, existing trackers struggle to capture the broader semantic context of target objects, including background elements and their relationships with the target \cite{vlt}.
Second, the lack of a temporal update mechanism for textual features over time further restricts the ability of existing trackers to perform long-term tracking effectively \cite{jointnlt}.
To address these issues, Guo \textit{et al.} proposed the target object attributes to provide high-level linguistic information and representation learning \cite{divert}. 
Li \textit{et al.} proposed the class name, color, texture, and material-based attributes for the target object using CLIP matching (Fig. \ref{fig2}(a)) \cite{citetracker}.
However, these studies ignore the contextual nuances and lack discriminative power in the presence of similar objects, leading to overfitting and reduced generalization \cite{citetracker}.

Textual descriptions play a crucial role in enhancing the performance of VLMs \cite{coop, coooop, waffle, dclip}. 
In the context of VOT, integrating language may provide valuable semantic guidance, particularly when visual cues are ambiguous or insufficient.
However, existing VLM-based trackers—such as CiteTracker \cite{citetracker}—often rely on a single, static phrase to describe each target object, which significantly limits their generalization capability, especially in complex or dynamic scenes \cite{tnl2k}.
These phrases are typically noun-based and lack detailed semantic and contextual (\(S\&C\)) information that can help disambiguate visually similar or occluded targets.
While recent trackers like CiteTracker \cite{citetracker} and JointNLT \cite{jointnlt} aim to enrich language inputs through attribute phrases or learnable prompts, they still operate within a narrow linguistic scope. 
Generative trackers attempt to produce or align more descriptive captions, but often do so in isolation and without modeling the temporal evolution of object appearances \cite{tnl2k, vlt, jointnlt, gti}.
As a result, these trackers struggle to adapt to real-world scenarios where target descriptions may vary based on context, viewpoint, or scene dynamics.

To overcome these limitations, we argue that trackers must move beyond fixed, shallow textual descriptions and instead utilize a comprehensive and context-aware language representation. Introducing richer and more varied textual descriptions—capturing not just class labels but also fine-grained attributes and scene context—may provide VLM-based trackers with a broader semantic grounding and improve their localization accuracy (Fig. \ref{fig1}). 
To our knowledge, no prior VLM-based tracker explicitly incorporates such diverse and temporally adaptive textual descriptions during either training or inference.

In this work, we propose CLDTracker, a novel vision-language tracker built upon a \textbf{C}omprehensive \textbf{L}anguage \textbf{D}escription (CLD) framework. 
CLDTracker is designed to address the limitations of prior methods by introducing two key innovations: a Bag of Textual Descriptions (\(B_t\)) and a Temporal Text Feature Update Mechanism (TTFUM) (Fig. \ref{fig2}(b)). 
The \(B_t\) module aggregates a diverse set of textual descriptions for each target—spanning class, attributes, and (\(S\&C\)) information—while TTFUM dynamically updates the relevance of these descriptions over time based on changes in the target’s visual appearance.
We define ``comprehensiveness'' as the joint modeling of multiple textual perspectives across different frames, enabling the tracker to associate an object with various ways it may be described in natural language (e.g., ``a bike in red,'' ``the runner,'' or ``the man near the car'').
By aligning multiple interrelated textual and visual representations, CLDTracker bridges the gap between static phrase-level grounding and the nuanced, temporally evolving nature of real-world tracking scenarios. 
This allows for more robust, adaptive, and semantically informed tracking, especially in cases where traditional one-to-one text–image alignment fails.

CLDTracker consists of two main branches, a text branch and a visual branch.
In the text branch, we generate a \(B_t\) by first compiling the predefined classes and attributes dictionaries cataloging various target objects using a range of publicly available datasets \cite{lasot, lasotext, trackingnet, tnl2k, lang_tracker, got10k}.
We then used an existing VLM \cite{clip} to select the most appropriate class name and attributes for each target object from this dictionary.
With GPT-4V \cite{gpt4v}, we transformed the first frame of the target object enclosed with a bounding box into a detailed textual description.
We then further enriched the target class name, attributes, and GPT-4V textual description into the \(S\&C\) using a \(S\&C\) enrichment module.
Using our extensive collection of textual descriptions from the visual content, we generated collections—or `bag'—of textual descriptions through an unsupervised and automated process. 
During pretraining, we learn textual prompts using our \(B_t\) for refining the comprehensive textual descriptions using adaptive and dynamic query matching.
These descriptions are then used to fine-tune our CLDTracker by correlating the model’s visual-textual embeddings to localize the target object.
Our model leverages comprehensive and well-structured textual descriptions to enhance VL tracking performance while reducing its sensitivity to variations in language input as shown in Fig. \ref{fig1}.
Since the target object may undergo appearance variation, therefore, we also introduce the TTFUM module within the text branch to update the target object's textual representation.
TTFUM uses the textual descriptions from previous frames instead of relying on the current frame to update the textual content.
By incorporating temporal features for the search region, the TTFUM module ensures precise comparison with the target region, and thus, CLDTracker improves object localization under target occlusion, out-of-view, and blurred motion.

In the vision branch, we extract visual features from the exemplar (representing the target object) and search images using a pretrained ViT architecture \cite{vit, mae}.
First, we concatenate the visual tokens from both images and pass them through the ViT, which serves as the backbone feature extractor.
To localize the target object's position, we estimate the correlation between textual features—derived from a comprehensive set of textual descriptions—and visual features.
The target object is identified based on the maximum value in the correlation map, and a scale estimate is performed using the prediction head \cite{stark}.

We performed extensive experiments on six publicly available benchmark datasets of VOT, including LaSOT \cite{lasot}, $\textrm{LaSOT}_{\textrm{ext}}$ \cite{lasotext}, TrackingNet \cite{trackingnet}, TNL2K \cite{tnl2k}, OTB99-Lang \cite{lang_tracker}, and GOT-10k \cite{got10k} and compared with 38 SOTA visual trackers.
Our results show significant performance improvements across six datasets compared to the SOTA trackers. 

In summary, our main contributions are as follows:

\begin{enumerate}
    \item We propose the concept of comprehensiveness in VOT. For this purpose, we construct a \(B_t\) containing textual descriptions in terms of class, attributes, and \(S\&C\) description of the target object (see Sec. \ref{sec:method}). To the best of our knowledge, no existing VLM-based tracker incorporates a comprehensive textual description of the target object (Sec. \ref{sec:method}). 
    \item We propose TTFUM within the textual branch of our CLDTracker. This module continuously update the textual representation of the target object in the temporal domain (Sec. \ref{sec:method}). 
    \item We perform rigorous experimental evaluations across six benchmark datasets and compared CLDTracker with 38 existing SOTA trackers (Sec. \ref{sec:results}).
\end{enumerate}

\noindent The rest of this paper is organized as follows: Sec. \ref{sec:relatedwork} reviews the SOTA VOT paradigms.
Sec. \ref{sec:method} presents our proposed CLDTracker in details.
Sec. \ref{sec:results} presents our rigorous experimental evaluation, while Sec. \ref{sec:conclusion} conclude this work and draws future directions.

\section{Related Work} \label{sec:relatedwork}
\noindent VOT has a long-established history in the literature, with numerous tracking paradigms proposed over time \cite{vot_survey, chen2022visual, trans_survey}.
In the following sections, we describe two major families of trackers: vision-only, which includes traditional and deep learning-based approaches \cite{vot_survey, trans_survey} that rely solely on visual information, and VLM-based methods \cite{jointnlt, citetracker, tnl2k, vlt, divert, UVLTrack}, which integrate the multi-modal representation to enhance VOT performance.

\subsection{Vision-only VOT}
\noindent Vision-only tracking has enjoyed great success in the past decades with numerous powerful paradigms have been emerged including DCFs-based \cite{kcf,dimp,prdimp, moose, atom, eco}, Siamese-based \cite{siamfc,siamattn,siamrpn,siamrpn++,siamban}, and transformer-based trackers \cite{transt,stark,ostrack,mixformer,mixformer2,seqtrack,artrack,aiatrack}.
The DCFs-based trackers learn the correlation filters online using the target template and then, they convolve it around the search region to estimate the maximum peak response \cite{kcf, moose,eco}.
End-to-end DCFs-based trackers are also proposed which learn the target object correlation filter in an offline manner such as in DiMP \cite{dimp}, ATOM \cite{atom}, and PrDiMP \cite{prdimp}, and then employed it for efficient target prediction.
DCFs-based trackers are fast however efficient target state estimation remains the major bottleneck in these trackers \cite{vot_survey}.

The Siamese tracking has emerged as a strong alternative to DCFs-based trackers by learning deep feature embeddings for powerful target representation and matching \cite{siamfc,siamattn,siamrpn,siamrpn++,siamban,vot_survey}.
The majority of the Siamese trackers consist of two identical branches that process the template and search regions simultaneously and compute similarities within a shared feature space \cite{siamfc,siamrpn}. 
SiamFC \cite{siamfc}, SiamRPN \cite{siamrpn}, and SiamRPN++ \cite{siamrpn++} trackers have demonstrated the potential of learning end-to-end features matching between the exemplar and the search regions.
However, Siamese trackers require large-scale and well-annotated image-paired datasets for offline training.
Moreover, existing Siamese trackers struggle to adapt to the target appearance variations during online tracking posing challenges in accurately estimating the target state \cite{vot_survey, siamese_survey}.

ViT architectures have emerged as a powerful paradigms in natural language processing \cite{bert} and computer vision applications \cite{vit, attention}.
ViT-based trackers have also been proposed, such as TransT \cite{transt}, STARK \cite{stark}, OSTrack \cite{ostrack}, MixFormer \cite{mixformer,mixformer2}, SeqTrack \cite{seqtrack}, AiATrack \cite{aiatrack}, DropTrack \cite{droptrack}, and SwinTrack \cite{swintrack}, among others.
The architectures of most of the ViT-based trackers follow the classical Siamese-based tracking paradigm with the only exceptions of plugging-in the ViT backbone as a powerful features extractor branch and efficient features fusion.
Though, ViT-driven trackers are leading the VOT performance leaderboards on almost all datasets compared to the DCFs and Siamese-based tracking paradigms \cite{trans_survey}, these trackers also required well annotated paired datasets, high computational cost due to the large model size, and poor adaptation to deformation an partial occlusion.

\subsection{Vision-Language-based VOT}
\noindent The integration of VLMs has opened new avenues, enabling visual trackers to incorporate semantic understanding and contextual reasoning \cite{lang_tracker,tnl2k,jointnlt,vlt,divert,citetracker,UVLTrack}.
Unlike vision-only trackers, VLM-based trackers also utilize textual representation in addition to the visual representation of the target object \cite{mmtrack,context_aware,vlm_survey}.
Both representations are aligned either using cross-modal fusion or cross-modal correlation modules for efficient target localization. 
The target state estimation component is normally borrowed from the classical trackers \cite{ostrack, transt}.

In the literature, various VLM-based trackers are recently proposed such as JointNLT \cite{jointnlt}, CiteTracker \cite{citetracker}, LangTracker \cite{lang_tracker}, ChatTracker \cite{chattracker}, TNL2K \cite{tnl2k}, VLT \cite{vlt,divert}, and QueryNLT \cite{context_aware}.
These SOTA trackers mainly consist of two main branches, including the vision and text branches.
Li \textit{et al.} proposed one of the first language-based trackers in which natural language representations are introduced within the VOT \cite{lang_tracker}.
Zhou \textit{et al.} recently proposed a JointNLT tracker in which ground truth textual description and visual representation of the target object are aligned within the ViT architecture \cite{jointnlt}.
Li \textit{et al.} proposed a CiteTracker in which the textual descriptions are enriched with a class name, color, texture, and material \cite{citetracker}.
Other recent VLM-based trackers include QueryNLT \cite{context_aware}, UVTrack \cite{UVLTrack}, and ChatTracker \cite{chattracker}.

\begin{figure*}[t!]
    \centering    
    \includegraphics[width=0.99\linewidth]{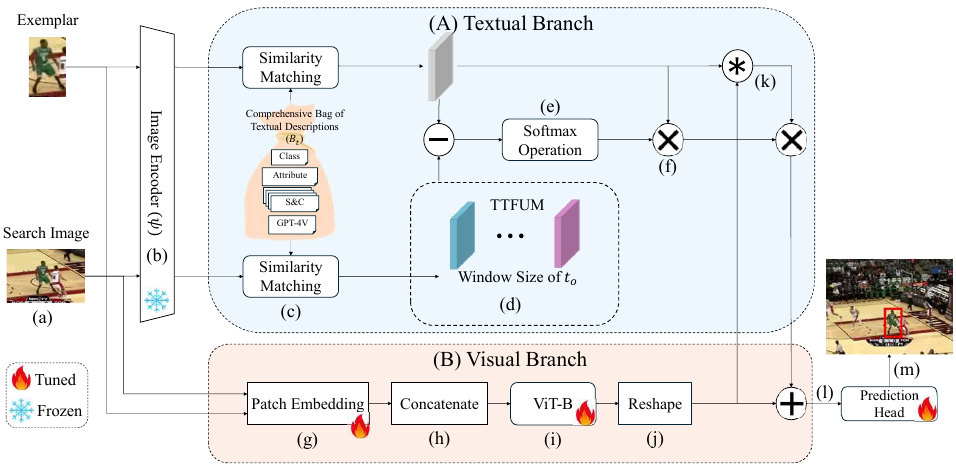}
    \caption{Schematic illustration of our proposed Comprehensive Language Description framework for Visual Tracking (CLDTracker). Our tracker comprises of two main branches including textual and visual. (A) shows our textual branch in which we first extract visual features from exemplar and search images using CLIP image encoder \cite{clip} (Steps (a) \& (b)), then we construct a \(B_t\) illustrated in Fig. \ref{fig:Bag} followed by the Prompt Adapter component and TTFUM in steps (c) \& (d). 
    (B) shows our vision branch in which we extract the visual features and then estimate the correlation with textual representation for efficient visual tracking.
    }
 \label{fig:CLDTracker}
 \end{figure*}

SOTA VLM-based trackers have demonstrated excellent performance in many VOT datasets.
However, these trackers also come up with their weaknesses.
For instance, existing trackers use a simple noun or phrase to represent a target object, which is not always sufficient as the target object may undergo appearance variations under different background scenarios.
CiteTracker addressed this issue. However, it shows performance degradation when the target object suffers from abrupt lighting conditions.
A limitation of these trackers is their potential inability to generalize well across different target objects due to the training dataset-specific biases consisting of paired textual-visual contents. 
Also, most of these VLM-based trackers primarily focus on aligning single textual and visual representations. 
In contrast, we propose the correlation of comprehensive visual-textual representation to concurrently align multiple correlated positive visual-textual concepts in our CLDTracker.
We argue that such an expansive and robust correlation significantly elevates performance across six VOT benchmark datasets.

\subsection{Advances in Foundation Models and Adaptation}
\noindent Recent developments in foundation models and cross-domain adaptation have greatly influenced multi-modal learning in vision tasks. UIU-Net \cite{reviewer_2} presents a unified framework for uncertainty-aware urban scene understanding, integrating deep learning and uncertainty modeling to enhance segmentation robustness. UrbanSAM \cite{reviewer_3} proposes invariance-inspired adapters for Segment Anything Models (SAM) \cite{sam}, showcasing powerful zero-shot generalization in urban construction scenarios. FlexiMo \cite{reviewer_4} introduces a flexible remote sensing foundation model that effectively adapts to diverse input modalities and label spaces, highlighting the strength of transfer learning and modular design in vision models.

While these models primarily target segmentation or remote sensing, their principles of modular adaptation, invariance, and prompt-based interaction align closely with the direction of our work. Specifically, our CLDTracker leverages comprehensive textual prompts and dynamic update mechanisms in a manner consistent with modular prompt tuning and adaptive model design seen in these works. Thus, we position CLDTracker as part of this broader movement toward generalizable, prompt-aware visual systems.

\section{Proposed CLDTracker} 
\label{sec:method}
\subsection{Overview} \label{sec:overview}
\noindent We propose the Comprehensive Language Description framework for visual Tracking (CLDTracker).
A system diagram of the proposed CLDTracker is shown in Fig. \ref{fig:CLDTracker}. 
CLDTracker effectively utilizes a collection of tracking sequences, paired with a predefined \(B_t\), to learn the correlation without any
ground truth textual descriptions.
The purpose is to tailor CLDTracker to a diverse range of VOT data gathered from various sources.
This enhances its generalization ability for adapting to unseen tracking sequences, especially for unfamiliar target object not encountered during the training phase.

\begin{figure*}[t!]
    \centering    
    \includegraphics[width=0.99\linewidth]{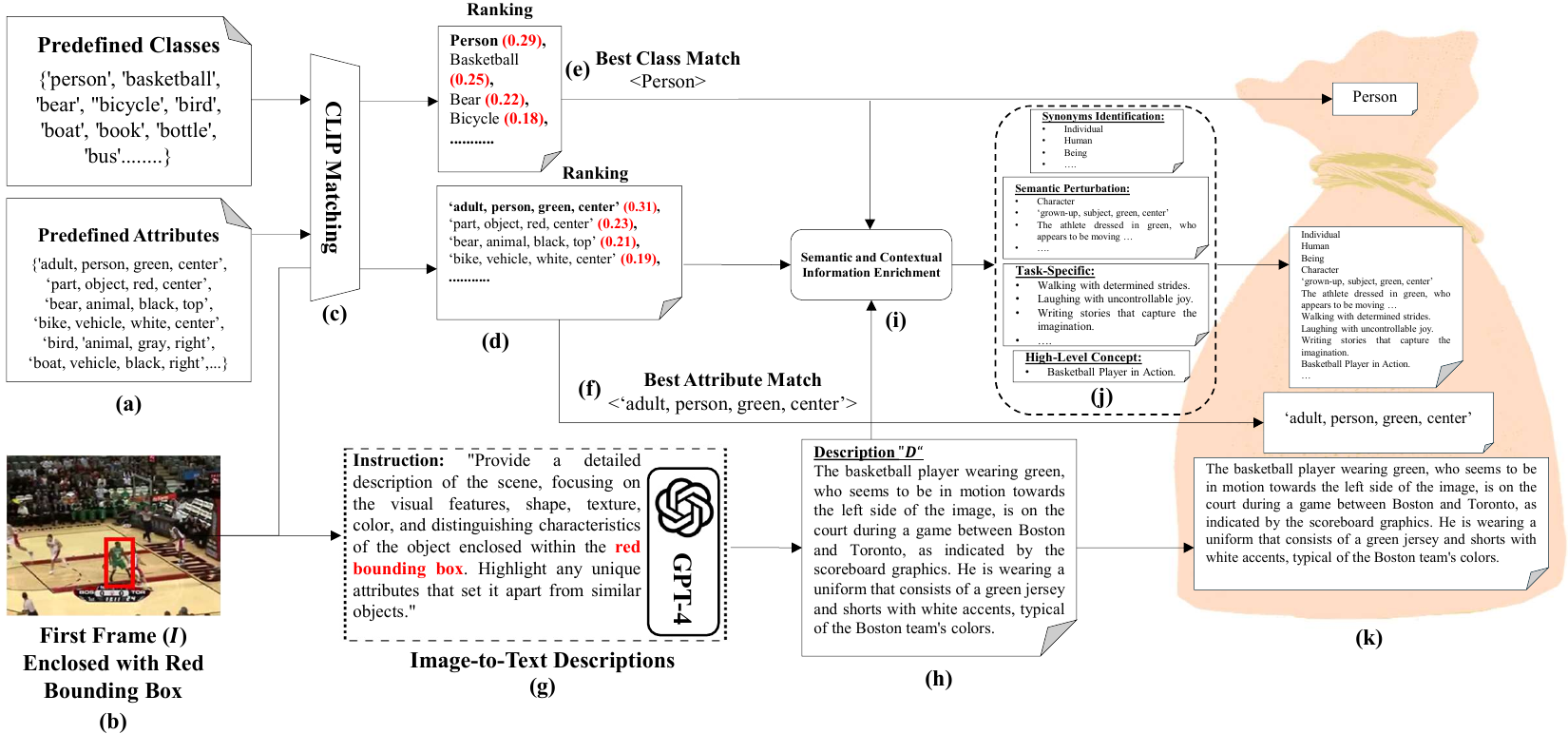}
    \caption{System diagram illustrating the process of building a \(B_t\). Steps (a) shows our class and corresponding attributes-based dictionaries. Step (b) shows the first frame of sequence containing target object tightly coupled with a bounding box. Steps (c)-(f) outline the process of identifying best class and attributes based on target object shown in step (b). Steps (g) \& (h) show the process of image-to-text descriptions using GPT-4V. Steps (i) \& (j) show the extension of the textual description by incorporating semantic and contextual information (\(S\&C\)), and step (k) shows an entire \(B_t\) containing overall descriptions.}
 \label{fig:Bag}
 \end{figure*}

CLDTracker mainly contains two branches: the textual and visual branches.
In the textual branch (Fig. \ref{fig:CLDTracker}(A)), we aim to construct a \(B_t\) of the target object by harnessing the existing SOTA VLMs \cite{clip, gpt4v} as shown in Fig. \ref{fig:Bag}.
It depicts the primary phases, including the construction of a predefined class and attributes dictionaries and the aggregation of corresponding textual descriptions, followed by the formation of \(S\&C\) contents.
These elements are integral to our correlation learning for target prediction, which seeks to align target object visual-textual pairs.
We then use the Prompt Adapter and TTFUM modules to refine and update the textual descriptions between the consecutive frames.
In the visual branch (Fig. \ref{fig:CLDTracker}(B)), we extract visual features from the exemplar and search images.
Both textual and visual representations are then utilized to pretrain the correlation loss in an end-to-end manners for efficient target localization.
The details of these processes are discussed in the following sections.

\subsection{Textual Branch (Fig. \ref{fig:CLDTracker}(A))}
\noindent The textual branch mainly generates a comprehensive textual descriptions of the target objects.
This is achieved by first compiling the target objects class names and corresponding target attributes followed by the construction of a \(B_t\).
We introduce Prompt Adapter module to enable adaptive and dynamic query matching for the selection of the most relevant description for each scenario from the \(B_t\).
Additionally, we develop our TTFUM within the textual branch to handle target appearance variations in the between the consecutive frames in the \(B_t\).

\subsubsection{Constructing a Comprehensive Bag of Textual Descriptions (Fig. \ref{fig:Bag})}
\noindent The system diagram of the \(B_t\) construction process is shown in Fig. \ref{fig:Bag} and Algorithm 1.
The process starts by first compiling the predefined classes and attributes dictionaries (Figs. \ref{fig:Bag}(a)-(b)) followed by the text expansion process by harnessing the existing VLMs (Figs. \ref{fig:Bag}(c)-(k)).
\\

\noindent \textbf{(a) Compiling Target Objects Class and Attributes Dictionaries:} The VOT datasets such as LaSOT \cite{lasot}, TNL2K \cite{tnl2k}, and OTB99-Lang \cite{lang_tracker}, among others are the only publicly accessible datasets for video-text pairing in VOT.
These datasets are used in SOTA trackers \cite{lang_tracker,gti,tnl2k,snlt,jointnlt,vlt} which restricts them to paired video-text data, that might not be comprehensively available to the public. 
To address this limitation, we propose a set target object class and attributes prompts dictionaries.
This serves as a foundational prompts to extract more comprehensive textual descriptions in subsequent phases.

We have created a strong dictionaries tailored to target object descriptions, which includes terms commonly used in natural language to describe various target objects, and their relevant attributes.
To generate this resource, we merged six existing datasets including LaSOT \cite{lasot}, $\textrm{LaSOT}_{\textrm{ext}}$ \cite{lasot}, TNL2K \cite{tnl2k}, OTB99-lang \cite{lang_tracker}, GOT-10k \cite{got10k}, and TrackingNet \cite{trackingnet}, and manually aggregated class names of each target to form a more precise target-specific dictionary.
In addition, we aggregated corresponding target object attributes from VLT \cite{divert} and formed target objects-specific attributes dictionary.
Our experiments compare the effectiveness of these two dictionaries, assessing the outcomes of each.

Our target-specific class and attributes dictionaries holds 940 and 23,899 varied class names and attributes, incorporating varying terms covering the range of target objects for VOT.
\textit{We have provided both dictionaries in our supplementary material and intend to release it to the public.}
\\

\noindent \textbf{(b) CLIP Matching for Identifying Best Class and Attributes (Figs. \ref{fig:Bag}(a)-(f)):} We define a set of classes \( C = \{ c_1, c_2, \ldots, c_M \} \) and corresponding target object attributes \cite{divert} \( A = \{ a_1, a_2, \ldots, a_N \} \), where $M$ and $N$ are the number of classes and attributes.
Given a first frame \( I \) of a sequence enclosed with a bounding box representing target object.
We aim to identify the best matching class and relevant attributes names of \( I \) by leveraging sets, \textit{C} and \textit{A} which are then utilized in the subsequent processing. 

This is achieved by first extracting the visual features \( F_{I} \) from \( I \) using the CLIP image encoder ($\Psi(\cdot;\theta)$) as: $F_{I} = \Psi(I) \in \mathbb{R}^q$, where $q$ denotes the features dimension.
The class and attributes textual features \(T_{C} \) and \( T_{A} \) are also extracted using the CLIP text encoder ($\Phi (\cdot;\phi)$):  $T_{C}= \Phi(C) \in \mathbb{R}^{M \times q}$ and $T_{A}= \Phi(A) \in \mathbb{R}^{N \times q}$. 
The best matched class and attributes are then identified using:

\begin{equation}
\hat{c}_{i}=\argmax_{c_{i} \in C} \textrm{sim} (F_{I}, T_{C_{i}}), \textrm{and},  
\hat{a}_{i}=\argmax_{a_{i} \in A} \textrm{sim} (F_{I}, T_{A_{i}}), 
    \label{eqn1}
\end{equation}

\noindent Here, $\textrm{sim}(F,T)=F^\top T/(||F||~||T||)$ denotes the cosine similarity measure.
Cosine similarity provides a normalized measure of alignment between two feature vectors, capturing their semantic relatedness while being invariant to magnitude differences. In this context, it evaluates how closely the extracted visual representation \(F_{I}\) aligns with the textual embeddings of candidate classes and attributes. Maximizing this similarity ensures that the selected class and attributes are those whose semantic meaning is most consistent with the visual content of the target object, facilitating robust and efficient matching within the shared feature space.

The resulting $\hat{c}_{i}$ and $\hat{a}_{i}$ are then added to the \(B_t\).
From here, we identify the top matching prompts in terms of class and attributes (Figs. \ref{fig:Bag}(e)-(f)).
Of these, only the best-matching class and attributes, termed \enquote{Person}, and \enquote{adult, Person, green, center}, are chosen and added to the \(B_t\) (Fig. \ref{fig:Bag}(k)). 
\\

\noindent \textbf{(c) GPT-4V for Comprehensive Textual Description (Fig. \ref{fig:Bag}(g)):} The image-to-text generative models have demonstrated excellent performance in the literature \cite{stablediffusion,vlm_survey}.
In this work, we derive the detailed caption of the target object by harnessing the existing LLM such as GPT-4V \cite{gpt4v} shown strong capabilities across numerous linguistics applications \cite{vlm_survey,awais2025foundation}.
To do so, we input first frame of the sequence enclosed with a bounding box and the following user prompt to the GPT-4V model as: ``\textit{Provide a detailed description of the scene, focusing on the visual features, shape, texture, color, and distinguishing characteristics of the object enclosed within the red bounding box. Highlight any unique attributes that set it apart from similar objects}.'' 
The model then returns the detailed description $D$ of the target object (Fig. \ref{fig:Bag}(h)) which is then added to the \(B_t\) (Fig. \ref{fig:Bag}(k)).

\noindent \textbf{(d) Semantic and Contextual Information (\(S\&C\)) (Figs. \ref{fig:Bag}(i)-(j)):} To further enrich the textual descriptions, we incorporate \(S\&C\) of the target object into the \(B_t\) \cite{waffle}.
This process consists of four main steps including synonyms identification, semantic perturbation, task-specific information, and high-level concept of the description.

Given a class word \( w \in C \), we define a retrieval function \( R(w) \) that retrieves a set of closely related synonyms \(w'\) for \( w \) class from a vocabulary set \( V \), where lexical diversity is introduced using WordNet \cite{bird2009natural}.
The \( R(w) \) is:
\begin{equation} \label{eq:syn_fnc}
    R(w) = \{ w' \in V \ | \ \text{sim}(w, w') \geq \tau \},
\end{equation}
where \( \tau \) is a threshold for semantic similarity.
In our case, we select the first $n=10$ words from \( R(w) \), yielding, $\mathcal{P}$ as:
\begin{equation} \label{eq:syn}
\mathcal{P} = \begin{cases} 
      w_1' \cdots w_n' & \text{if } R(w) \neq \emptyset \\
      w & \text{otherwise.} 
   \end{cases}
\end{equation}

In the second step, we introduce diversity while preserving the original semantic meaning in a bag \(\mathcal{B_i}\). 
This is achieved by adding semantic perturbation to each token \( t_i \) in a bag \( \mathcal{B_t} = \{ t_1, t_2, \ldots, t_n \} \). We define a perturbation strength parameter \( \alpha \in [0,1] \), which controls the probability of replacing each token \( t_i \) with a synonym. The replacement probability \( p_i \) is formally defined as:
\begin{equation}
    p_i = \begin{cases} 
      1 & \text{if } \text{rand()} < \alpha \\
      0 & \text{otherwise,} 
   \end{cases}
\end{equation}
\noindent where \( \text{rand()} \) generates a random value in \([0,1]\).
The perturbed sequence \( B_{\text{pert}} = \{ t_1', t_2', \ldots, t_n' \} \) is formed by selecting replaced tokens in the original \(B_t\) based on the perturbation probability which introduces diversity while preserving the original semantic meaning in the original \(B_t\).

In the third step, we introduce task-specific information of the target object harnessing the GPT-4V model.
This is achieved by using the following prompt in the GPT-4V model as: ``\textit{Create a 10 unique and detailed descriptive phrases or scenarios within a \{\( c_{i} \in C \)\}. Each should vividly capture distinctive actions, characteristics, or typical contexts, emphasizing variety and creativity while maintaining relevance to the theme.''}.
The GPT-4V outputs the following set of phrases  \( \mathcal{N}_c \) for class $c_{i}=``\textrm{\textit{Bird}}$'' as:
\[
\mathcal{N}_{\text{Bird}} = \left\{ 
\begin{array}{l}
\text{``soaring through the sky",} \\
\text{``singing a beautiful melody",} \\
\ldots 
\end{array} 
\right\}
\]

\noindent A semantic information corresponding to the target object is aggregated by concatenating the $B_{pert}$ and $N_{\text{task}}$:
\begin{equation}
    S = \text{Concat}(B_{\text{pert}}, N_{\text{task}})
\end{equation}

In the final step, we incorporate a high-level concept using the target object description $D$ harnessing GPT-4V model using the following input prompt:
``\textit{Based on the description $D$, provide a concise class name (in five words or fewer) that best represents its category.}''
All these \(S\&C\) is then finally added to the \(B_t\).

\noindent \textbf{(e) Comprehensive Bag of Textual Descriptions Validation and Enhancement Pipeline:}
To encourage concise and visually grounded descriptions, we apply prompt engineering techniques with carefully crafted templates and few-shot examples. Each description \(D_i \in B_{\text{raw}}\) is encoded using the CLIP text encoder, producing \(d_i = \Phi(D_i)\), and evaluated against the visual feature embedding \(F_I\) via cosine similarity \(\textrm{sim}(F_I, d_i) = \frac{F_I^\top d_i}{\|F_I\| \cdot \|d_i\|}\).
Descriptions that meet the semantic alignment constraint \(\textrm{sim}(F_I, d_i) \geq \tau, \quad \text{with} \quad \tau = 0.8\), are retained in a filtered set \(B_{\text{valid}} = \{D_i \in B_{\text{raw}} \mid \textrm{sim}(F_I, \Phi(D_i)) \geq \tau\}\). Those that fail this threshold are discarded and regenerated.
The valid descriptions are then manually reviewed for correctness and consistency, resulting in a refined set \(B_{\text{final}}\). Finally, each entry in \(B_{\text{final}}\) is encoded using the CLIP text encoder, yielding the final bag of textual feature embeddings \(B_t = \Phi(B_{\text{final}}) = \{\Phi(D_i) \mid D_i \in B_{\text{final}}\}\).
This pre-encoding strategy avoids redundant computation during training and inference, significantly improving runtime efficiency. Further details are provided in Appendix Sec. \ref{sec:bag_complexity}.

\subsubsection{Prompt Adapter of Textual Descriptions}
\noindent This module generates an adaptive textual representation of the target object to select the most relevant description from \(B_t\) based on the target's visual features.
The \(B_t\) may be noisy or contains irrelevant description of the target object.
Therefore, instead of utilizing the entire \(B_t\), we learn the prompts for the textual description using SOTA Conditional Context Optimization (CoCoOp) that employs instance-conditioned prompts learning mechanism to dynamically refine Vision-Language (VL) alignment \cite{coooop}.
CoCoOp consists of two learnable components including a set of context vectors and a lightweight neural network that generates for each image an input-conditional token.

Let $d_i$ be the $i$-th textual embedding from \(B_t\) and \{$\boldsymbol{\upsilon}_1, \boldsymbol{\upsilon}_2, \ldots, \boldsymbol{\upsilon}_L$\} be the $L$ learnable context vectors, where each $\boldsymbol{\upsilon}_i$ has the same dimensions as $d_i$.
These context vectors act like learnable "prefixes" or "hints" that condition each textual description to better match the visual features of the current target image. This allows the model to generate more context-aware text embeddings.
Our objective is to compute the cosine similarity between the visual features (\(F_{e}\)) corresponding to the exemplar image (\(I_{e}\)) and the textual prompts $t_i(F_{e}) = \{\boldsymbol{\upsilon}_1(F_{e}), \boldsymbol{\upsilon}_2(F_{e}), \ldots, \boldsymbol{\upsilon}_L(F_{e}), d_i\}$ as:
\begin{equation} \label{eq:coooop}
    p(d_i \mid F_{e}) = \frac{\exp(\text{sim}(F_{e}, \Phi(t_i(F_{e}))) / \tau)}{\sum_{j=1}^K \exp(\text{sim}(F_{e}, \Phi(t_j(F_{e}))) / \tau)}
\end{equation}
where $\tau$ is a learnable temperature parameter, and $K$ is the number of descriptions in a \(B_t\).
The target description of the exemplar image \(T_{e}\) is then predicted using $T_{e}(index) = T_{\text{index}}$, where index is estimated from \(p(d_i \mid F_{e})\) using: $\text{index} = \arg\max(p_i), \, i \in (1, K)$.
Similarly, for the search image we get \(T_{s}\) using Eq. (\ref{eq:coooop}).
Eq.~(\ref{eq:coooop}) dynamically scores how well each text description matches the visual features of the target object, after adapting the descriptions through learned context vectors. By focusing on the highest scoring text, the model robustly selects the most semantically relevant and instance-specific descriptions, filtering out irrelevant or noisy candidates.
Finally, the normalized adaptive text representations for both \(T_{e}\) and \(T_{s}\) is then obtained using the softmax function as \cite{citetracker}:
\begin{equation} \label{eq:proj}
    \tilde{T}_{e} = \text{Softmax}(W_{\text{proj}} \cdot T_{e}),
\end{equation}
\noindent where \(W_{proj}\) is the weighting matrix of the linear layer which is used to align the dimension with the visual features.
Similarly, we get normalized textual representation $\tilde{T}_{s}$ for the search image using Eq. (\ref{eq:proj}).
This projection step ensures that the selected textual embeddings are dimensionally compatible with the visual feature space, enabling effective fusion of visual and textual cues in the subsequent processing stages.

\subsubsection{Temporal Text Feature Update Mechanism (Fig. \ref{fig:CLDTracker}(d))}
\noindent The target object in the video sequence may undergo abrupt appearance variations; therefore, the textual descriptions are required to adapt to the background variations.
For this purpose, we introduce our TTFUM, which addresses this issue by updating textual feature embeddings of the target object across frames.

We first estimate the average textual features that capture the information of the past $n$ frames.
Then, these features are used to compute the absolute difference with the exemplar embedding.
Finally, we compute the attention weights using:

\begin{equation} \label{eq:ttfum}
    W_{att} = \textrm{softmax}\Bigg(- \Big|\tilde{T}_{e} - \frac{1}{n} \sum_{i=t-n+1}^{t} \tilde{T}_{s}(i)\Big|\Bigg),
\end{equation}

\noindent which are then aggregated with the exemplar textual features, resulting in a  dynamic textual features update as: $T_{att} = W_{att} \cdot \tilde{T}_{e}$.

\begin{algorithm} \label{algo}
\caption{\(B_t\) construction}
\begin{algorithmic}[1]
\Require Input image target $I$
\State \textbf{Embedding}:
    \State \quad $F_{I} = \Psi(I)$
    \State \quad $F_{d_{class}} = \Phi(C)$
    \State \quad $F_{d_{attributes}} = \Phi(A)$
\State \textbf{Matching}:
    \State \quad $(j^*, k^*) = \arg\max_{j,k} \Big( \text{sim}(F_{I}, F_{d_{class,j}})$ 
\State \quad \quad \quad \quad $+ \text{sim}(F_{I}, F_{d_{attributes,k}}) \Big)$
\State \textbf{Description Generation}:
    \State \quad $D = \text{GPT-4V}(I)$
\State \textbf{\(S\&C\) Enrichment Module}:
    \State \quad $S\&C = \text{ConstructSequence}(C_{matched}, A_{matched}, D)$
\State \textbf{Output}:
    \State \quad $B_t = \{ C_{matched}, A_{matched}, D, S\&C \}$
\end{algorithmic}
\end{algorithm}

\subsection{Visual Branch (Fig. \ref{fig:CLDTracker}(B))}
\noindent CLDTracker is a multi-modal representation learning paradigm that integrates the textual features estimated using the text branch with the visual feature representation of the search image ($I_{s}$) and exemplar image ($I_{e}$).
For this purpose, we extract visual features using the ViT-B architecture \cite{vit} and compute the correlation.

We borrow the visual branch from the OSTrack \cite{ostrack} that has already been used by many SOTA trackers \cite{citetracker, droptrack, divert}.
We first divide the search and exemplar regions into small patches and consider them as a visual token \cite{vit}.
We concatenate the flattened exemplar and search regions visual tokens and then feed them into staked multi-self attention layers vision encoder to get the visual features of the current frame \( F_{\text{frame}} \in \mathbb{R}^{D} \).
The self-attention module within the vision encoder captures the feature variations iteratively between the $I_{s}$ and $I_{e}$ regions.
This allows mutual guidance for target oriented feature extraction (Fig. \ref{fig:CLDTracker}(B)(g-j)).
The resulting search region features \( F_{\text{frame}} \in \mathbb{R}^{D} \) are reshaped which are then directly used for target classification and regression with the textual representations.

\subsection{Target Prediction Using Image-Bag Correlation}
\noindent We estimate the correlation $Corr(F_{\text{frame}}, \tilde{T}_e)$ between the textual features \( \tilde{T}_e \) and the visual features \( F_{\text{frame}} \) using a convolution operation in which text features are used as the kernel weights \cite{citetracker}.
Mathematically, it is represented as:
\begin{equation} \label{eq:corr}
\textrm{Corr}(F_{\text{frame}}, \tilde{T}_e) =
(1 + T_{att} (\tilde{T}_e) ) * F_{\text{frame}},
\end{equation}
\noindent where \(*\) represents the convolution operation.

\subsection{Target State Estimation}
\noindent The correlated features $\textrm{Corr}(F_{\text{frame}}, \tilde{T}_e)$ estimated in Eq. (\ref{eq:corr}) are used to predict the target state. 
We used a commonly used prediction head to estimate the target state \cite{ostrack} which consists of four stacked Conv-BN-ReLU layers. 
This target state prediction head outputs a classification score map C, offset maps O for compensating for reduced resolution, and size maps S.
The target state $(x,y,w,h)$ is computed as:

\begin{equation}
 (x,y,w,h)= (x_{c}+O_{x},y_{c}+O_{y},S_{w},S_{h}),
    \label{eqn:state}
\end{equation}

\noindent where $(x_{c},y_{c})$ is the target centre estimated as: $(x_{c},y_{c})= argmax_(x,y)C(xy)$.
$(O_{x},O_{y})$ represents the shifts to $(x_{c},y_{c})$ from O.
$(S_{w},S_{h})$ is the predicted box size from $S$.

\subsection{Overall Training Loss}
\noindent We employ commonly used loss functions including classification and regression \cite{ostrack,citetracker} to train the proposed CLDTracker.
The classification loss is based on weighted focal loss \cite{cornernet} \(L_{\text{cls}}\).
For bounding box regression, we utilized two losses including \( L_1 \) loss and Intersection over Union (IoU) loss \(L_{\text{iou}}\) \cite{giou_loss}. 
The total loss function is defined as:
\begin{equation} \label{eq:loss}
    L_{\text{track}} = L_{\text{cls}} + \lambda_{\text{iou}} L_{\text{iou}} + \lambda_{L_1} L_1,
\end{equation}
The regularization parameters \( \lambda_{\text{iou}} = 2 \) and \( \lambda_{L_1} = 5 \) are set empirically in our experiments.

\subsection{Inference}
\noindent During the inference phase, a Hanning window penalty is applied to incorporate positional priors and refine the confidence scores. Specifically, the prediction head outputs a set of 1,024 bounding boxes, each associated with a confidence score, and the Hanning window with a shape of \( 32 \times 32 \) is applied to these scores for post-processing. This window penalizes feature points that are far from the target, helping to refine the localization of the object within the search region.

The final confidence score \( \text{score}_w \) is computed by a weighted combination of the original score \( \text{score} \) and the Hanning window score \( \text{score}_h \), using the following formula:
\begin{equation} \label{eq:infernece}
    \text{score}_w = (1 - w) \times \text{score} + w \times \text{score}_h,
\end{equation}
where \( w \) is a weighting parameter chosen empirically, with \( w = 0.49 \) in this case. The variable \( \text{score}_h \) corresponds to the value of the Hanning window at the same spatial position as the original score.
By applying this penalty, the tracker reduces the confidence of bounding boxes that are positioned far from the target based on prior frames. After this post-processing step, the bounding box with the highest final confidence score \( \text{score}_w \) is selected as the final tracking result.

\section{Experiments} \label{sec:results}
\noindent We conducted extensive experiments to evaluate the performance of our proposed CLDTracker across six publicly available VOT benchmark datasets including LaSOT \cite{lasot}, $\textrm{LaSOT}_{ext}$ \cite{lasotext}, TrackingNet \cite{trackingnet}, GOT-10k \cite{got10k}, TNL2K \cite{tnl2k}, and OTB99-Lang \cite{lang_tracker}.
We also compared the performance with 38 existing SOTA visual and VLM-based trackers. 
These experiments, covering a range of tracking challenges, allowed us to thoroughly assess CLDTracker performance. 
We begin by explaining the training and implementation details, the VOT datasets, the evaluation metrics, and the SOTA trackers for comparison.
We then present the qualitative and quantitative evaluations, followed by the ablation studies and computational time.

\subsection{Training and Implementation Details}
\noindent Our CLDTracker is implemented using a workstation with 4 NVIDIA A100 GPUs, completing the full training process in approximately 27 hours, which corresponds to a total compute cost of 108 GPU hours.
The inference speed is evaluated on a single NVIDIA GeForce RTX 3080 GPU.
The textual branch leverages CLIP-ViT-B/32 \cite{clip}, where both the image and text encoders are initialized from the pretrained CLIP model.
For the visual branch, we use a ViT-B model~\cite{vit} pretrained with the MAE method~\cite{mae} as the backbone feature extractor for joint visual feature extraction and relational modeling.
The state prediction head consists of a lightweight FCN, consisting of 4 stacked \texttt{Conv-BN-ReLU} layers.
The search image $I_s$ is cropped four times the area of the bounding box in the test frame and resized to $384 \times 384$ pixels.
Similarly, the exemplar image $I_e$ is cropped twice the target box area and resized to $192 \times 192$ pixels. 

We trained CLDTracker following standard VOT practice~\cite{stark, ostrack, droptrack, citetracker}, using the training splits of GOT-10k \cite{got10k}, TrackingNet \cite{trackingnet}, and LaSOT \cite{lasot}. 
In addition, we also utilized COCO2017 \cite{coco} with using only the official textual prompts and class names.
We use standard data augmentation schemes, including horizontal flips, color saturation, and jittering during training. 
CLDTracker is trained using a batch size of 128 with AdamW optimizer \cite{adamw}.
We use  a weight decay of \(10^{-4}\) and the initial learning rate is set to \(4 \times 10^{-5}\).
We used 300 epochs, with 60,000 image pairs per epoch, and the learning rate is reduced by a factor of 10 after 240 epochs.

\subsection{VOT Datasets}
\noindent We used six publicly available VOT datasets which are summarized as follows:
\begin{itemize}

    \item \textbf{LaSOT \cite{lasot}}: is a large-scale dataset comprising 1,120 long-term videos for training and 280 for testing.
    Each video is annotated with high-quality bounding boxes. The videos have an average length of 2,448 frames per sequence.
    Additionally, each video is accompanied by a natural language description.

    \item \textbf{LaSOT\textsubscript{ext} \cite{lasotext}:} expands the LaSOT \cite{lasot} dataset by introducing 150 challenging test videos spanning 15 object classes. 
    These videos feature diverse tracking challenges, such as occlusions and small object variations. 
    Unlike the original LaSOT, this dataset does not include natural language descriptions for the video sequences.
    
    \item \textbf{TrackingNet \cite{trackingnet}:} is a large-scale tracking dataset designed to encompass a diverse range of object classes and real-world scenarios. 
    Its test set consists of 511 sequences, with only the initial frame annotated. The dataset does not include natural language descriptions for the sequences.
    We evaluated our tracker on the TrackingNet test set and submitted the results to the official evaluation server for benchmarking.
 
    \item \textbf{TNL2k \cite{tnl2k}:} is a large-scale VL benchmark designed for tracking via natural language descriptions.
    It comprises 2,000 video sequences, with a training/testing split of 1,300/700. 
    Each video is annotated with a descriptive sentence and a corresponding bounding box that specifies the target object and its precise location.

    \item \textbf{OTB99-Lang \cite{lang_tracker}}: dataset incorporates sentence descriptions for target objects in each video. 
    This benchmark comprises 99 challenging sequences, with a training/testing split of 51/48. This dataset also provides a diverse and well-balanced evaluation framework for VL tracking.
    
    \item \textbf{GOT-10k \cite{got10k}:} is a large-scale dataset comprising over 10k videos for training and 180 videos for testing without any associated natural language descriptions.
    It maintains a strict zero-overlap policy between object classes in the training and testing subsets.
    Following the official protocols, which prohibits the use of external datasets for training, we evaluated our model under similar conditions.

    \item \textbf{VastTrack \cite{vasttrack}:} A large-scale dataset comprising over 47k training videos and 3.5k testing videos, each paired with natural language descriptions. It follows a hybrid evaluation protocol, where some object classes in the test set overlap with the training set, while the remaining classes are entirely unseen. VastTrack spans 2,115 object categories, significantly more than those in existing popular benchmarks \cite{lasot, got10k}.
    
\end{itemize}

\subsection{Evaluation Metrics}
\noindent Following the official VOT experimental protocols defined by LaSOT \cite{lasot}, we use Success (S), Precision (P), and Normalized Precision (NP) to evaluate the performance of the trackers.
We used One-Pass Evaluation (OPE) and rank the existing SOTA trackers according to the evaluation metrics. 
For GOT-10k \cite{got10k} dataset, we employed Average Overlap (AO) and Success Rate (SR) using a threshold of 0.50 and 0.75.

\subsection{SOTA Trackers for Comparison}
\noindent We compared the performance of our proposed CLDTracker with 38 existing SOTA trackers. For fairness, we leave all default configurations, training settings, and hyperparameters of the compared methods untouched. Evaluation metrics are either recomputed using publicly available raw predictions or taken directly from the authors’ original publications and official source codes when predictions are not available.
We categorize the SOTA trackers into three categories including 6 CNN-based trackers, 20 ViT-based trackers, and 12 VLM-based trackers.

The CNN-based trackers include ECO \cite{eco}, ATOM \cite{atom}, DiMP50 \cite{dimp}, PrDiMP50 \cite{prdimp}, Ocean \cite{ocean}, and KeepTrack \cite{keeptrack}.
The ViT-based trackers include TransT \cite{transt}, STARK-ST50 \cite{stark}, TrDiMP \cite{trdimp}, UTT \cite{utt}, ToMP50 \cite{tomp}, CSWinTT \cite{cswintt}, AiATrack \cite{aiatrack}, MixFormer-22k \cite{mixformer}, SwinTrack-B-384 \cite{swintrack}, OSTrack-384 \cite{ostrack}, GRM \cite{grm}, VideoTrack \cite{videotrack}, SeqTrack-B256 \cite{seqtrack}, DropTrack \cite{droptrack}, MAT \cite{mat}, ARTrack\textsubscript{384} \cite{artrack}, MixViT (ConvMAE) \cite{mixformer}, DiffusionTrack-B256 \cite{DiffusionTrack}, HIPTrack \cite{hiptrack}, and AQATrack-256 \cite{AQATrack}.
While the VLM-based trackers include RTTNLD \cite{feng}, GTI \cite{gti}, TNL2K-2 \cite{tnl2k}, SNLT \cite{snlt}, Li et al, \cite{cmtr}, VLT\textsubscript{TT} \cite{vlt}, CiteTracker \cite{citetracker}, JointNLT \cite{jointnlt}, VLT\textsubscript{OST-384} \cite{divert}, UVLTrack-B \cite{UVLTrack}, QueryNLT \cite{context_aware}, and MMTrack \cite{mmtrack}.

\begin{table*}
\centering
\caption{Comparison of our method with SOTA methods on LaSOT, LaSOT\textsubscript{ext}, TrackingNet, TNL2K, OTB99-Lang, and GOT-10k datasets. The best two results are shown in \textcolor{red}{red}, and \textcolor{blue}{blue} color, respectively}

\resizebox{\textwidth}{!}{

\begin{tabular}{lll|ccc|ccc|ccc|ccc|ccc|ccc}
 
\specialrule{2pt}{0pt}{0pt}
& \multirow{2}{4em}{Method} & \multirow{2}{4em}{Source} & \multicolumn{3}{|c|}{LaSOT}& \multicolumn{3}{|c|}{LaSOT\textsubscript{ext}}& \multicolumn{3}{|c|}{TrackingNet}& \multicolumn{3}{|c|}{TNL2K}&\multicolumn{3}{c}{OTB99-Lang}& \multicolumn{3}{|c}{GOT-10k} \\
\cline{4-21}

& & & S & NP & P & S & NP & P & S & NP & P & S & NP & P & S & NP & P & AO &  SR\textsubscript{0.50} & SR\textsubscript{0.75} \\

\specialrule{2pt}{0pt}{0pt}

\parbox[t]{0.01cm}{\multirow{6}{*}{\rotatebox[origin=c]{90}{\textbf{CNN}}}}

& ECO \cite{eco} & CVPR17
& 32.4 & - & 30.1 
& 22.0 & - & 24.0 
& 55.4 & - & 49.2 
& - & - & - 
& - & - & - 
& 31.6 & 30.9 & 11.1 
\\

& ATOM \cite{atom} & CVPR19
& 51.5 & 57.6 & 50.5 
& 37.6 & 45.9 & 43.0 
& 70.3 & 77.1 & 64.8 
& 40.0 & 47.0 & 39.0 
& 67.6 & - & 82.4 
& 55.6 & 63.4 & 40.2 
\\

& DiMP50 \cite{dimp} & ICCV19
& 56.9 & 65.0 & 56.7 
& 39.2 & 47.6 & 45.1 
& 74.0 & 80.1 & 68.7 
& 44.7 & - & 43.4 
& - & - & - 
& 61.1 & 71.7 & 49.2 
\\

& PrDiMP50 \cite{prdimp} & CVPR20
& 59.8 & - & 60.8 
& - & - & - 
& 75.8 & 81.6 & 70.4 
& 47.0 & - & 45.9 
& 69.5 & - & 89.5 
& 63.4 & 73.8 & 54.3 
\\

& Ocean \cite{ocean} & ECCV20
& 56.0 & 65.1 & 56.6 
& - & - & - 
& - & - & - 
& 38.4 & - & 37.7 
& - & - & - 
& 61.1 & 72.1 & 47.3 
\\

& KeepTrack \cite{keeptrack} & ICCV21
& 67.1 & 77.2 & 70.2 
& 48.2 & 58.0 & - 
& - & - & - 
& - & - & - 
& - & - & - 
& - & - & - 
\\

\specialrule{2pt}{0pt}{0pt}
 
\parbox[t]{0.01cm}{\multirow{20}{*}{\rotatebox[origin=c]{90}{\begin{tabular}[l]{@{}l@{}}\textbf{ViT-Based}\end{tabular}}}}

& TransT \cite{transt} & CVPR21
& 64.9 & 73.8 & 69.0 
& 44.8 & - & 52.5 
& 81.4 & 86.7 & 80.3 
& 50.7 & - & 51.7 
& 70.8 & - & 91.2 
& 67.1 & 76.8 & 60.9 
\\

& STARK-ST50 \cite{stark} & ICCV21
& 66.4 & - & 71.2 
& 47.8 & - & 55.1 
& 81.3 & 86.1 & - 
& - & - & - 
& 69.6 & - & 91.4 
& 68.0 & 77.7 & 62.3 
\\

& TrDiMP \cite{trdimp} & CVPR21
& 63.9 & - & 66.3 
& - & - & - 
& 78.4 & 83.3 & 73.1 
& - & - & - 
& - & - & - 
& 67.1 & 77.7 & 58.3 
\\

& UTT \cite{utt} & CVPR22
& 64.6 & - & 67.2 
& - & - & - 
& 79.7 & - & 77.0 
& - & - & - 
& - & - & - 
& 67.2 & 76.3 & 60.5 
\\

& ToMP50 \cite{tomp} & CVPR22
& 67.6 & 78.0 & 72.2 
& 45.4 & - & - 
& 81.2 & 86.2 & 78.6 
& - & - & - 
& - & - & - 
& 69.6 & 80.0 & 63.2 
\\

& CSWinTT \cite{cswintt} & CVPR22
& 66.2 & 75.2 & 70.9 
& - & - & - 
& 81.9 & 86.7 & 79.5 
& - & - & - 
& - & - & - 
& 69.4 & 78.9 & 65.4 
\\

& AiATrack \cite{aiatrack} & ECCV22
& 69.0 & 79.4 & 73.8 
& 46.8 & 54.4 & 54.2 
& 82.7 & 87.8 & 80.4 
& - & - & - 
& - & - & - 
& 69.6 & 80.0 & 63.2 
\\ 
    
& MixFormer-22k \cite{mixformer} & CVPR22
& 69.2 & 78.7 & 74.7 
& - & - & - 
& 83.1 & 88.1 & 81.6 
& - & - & - 
& 71.0 & - & 93.1 
& 72.6 & 82.2 & 68.8 
\\

& SwinTrack-B-384 \cite{swintrack} & NeurIPS22
& 71.3 & - & 76.5 
& 49.1 & - & 55.6 
& 84.0 & - & 82.8 
& 55.9 & - & 57.1 
& - & - & - 
& 72.4 & 80.5 & 67.8 
\\

& OSTrack-384 \cite{ostrack} & ECCV22 
& 71.1 & 81.1 & 77.6 
& 50.5 & 61.3 & 57.6 
& 83.9 & 88.5 & 83.2 
& 55.9 & - & 56.7 
& 69.6 & \textcolor{blue}{\textbf{85.1}} & 92.2 
& 73.7 & 83.2 & 70.8 
\\

& GRM \cite{grm} & CVPR23
& 69.9 & 79.3 & 75.8 
& & & 
& 84.0 & 88.7 & 83.3 
& - & - & - 
& - & - & - 
& 73.4 & 82.9 & 70.4 
\\

& VideoTrack \cite{videotrack} & CVPR23
& 70.2 & - & 76.4 
& - & - & - 
& 83.8 & 88.7 & 83.1 
& - & - & - 
& - & - & - 
& 72.9 & 81.9 & 69.8 
\\

& SeqTrack-B256 \cite{seqtrack} & CVPR23
& 69.9 & 79.7 & 76.3 
& 49.5 & 60.8 & 56.3 
& 83.3 & 88.3 & 82.2 
& 54.9 & - & - 
& - & - & - 
& 74.7 & 84.7 & 71.8 
\\

& DropTrack \cite{droptrack} & CVPR23 
& 71.8 & 81.8 & 78.1 
& 52.7 & 63.9 & 60.2 
& 84.1 & 88.9 & - 
& 56.9 & - & 57.9 
& - & - & - 
& 75.9 & 86.8 & 72.0 
\\

& MAT \cite{mat} & CVPR23
& 67.8 & 77.3 & - 
& - & - & - 
& 81.9 & 86.8 & - 
& 51.3 & - & - 
& - & - & - 
& 67.7 & 78.4 & - 
\\

& ARTrack\textsubscript{384} \cite{artrack} & CVPR23
& 72.6 & 81.7 & 79.1 
& 51.9 & 62.0 & 58.5 
& \textcolor{red}{\textbf{85.1}} & \textcolor{blue}{\textbf{89.1}} & \textcolor{blue}{\textbf{84.8}} 
& 59.8 & - & - 
& - & - & - 
& 75.5 & 84.3 & 74.3 
\\

& MixViT (ConvMAE) \cite{mixformer2} & TPAMI24
& 70.4 & 80.4 & 76.7 
& - & - & - 
& \textcolor{blue}{\textbf{84.5}} & \textcolor{blue}{\textbf{89.1}} & 83.7 
& - & - & - 
& - & - & - 
& 74.3 & 84.1 & 73.0 
\\

& DiffusionTrack-B256 \cite{DiffusionTrack} & CVPR24
& 70.8 & 79.8 & 76.7 
& - & - & - 
& 83.8 & 88.2 & 82.1 
& 56.4 & 72.5 & 57.3 
& - & - & - 
& 74.8 & 85.4 & 72.0 
\\

& HIPTrack \cite{hiptrack} & CVPR24
& \textcolor{blue}{\textbf{72.7}} & \textcolor{blue}{\textbf{82.9}} & \textcolor{blue}{\textbf{79.5}} 
& 53.0 & \textcolor{blue}{\textbf{64.3}} & \textcolor{blue}{\textbf{60.6}} 
& \textcolor{blue}{\textbf{84.5}} & \textcolor{blue}{\textbf{89.1}} & 83.8 
& - & - & - 
& - & - & - 
& \textcolor{blue}{\textbf{77.4}} & \textcolor{red}{\textbf{88.0}} & \textcolor{blue}{\textbf{74.5}} 
\\

& AQATrack-256 \cite{AQATrack} & CVPR24
& 71.4 & 81.9 & 78.6 
& 51.2 & 62.2 & 58.9 
& 83.8 & 88.6 & 83.1 
& 57.8 & 59.4 & - 
& - & - & - 
& 73.8 & 83.2 & 72.1 
\\

\specialrule{2pt}{0pt}{0pt}

\parbox[t]{0.01cm}{\multirow{12}{*}{\rotatebox[origin=c]{90}{\textbf{VLM}}}}

& RTTNLD \cite{feng} & WACV20
& 35.0 & 43.0 & 35.0 
& - & - & - 
& - & - & - 
& - & - & - 
& 61.0 & 73.0 & 79.0 
& - & - & - 
\\

& GTI \cite{gti} & TCSVT21
& 63.1 & - & 66.5 
& - & - & - 
& - & - & - 
& - & - & - 
& 67.2 & - & 86.3 
& - & - & - 
\\

& TNL2K-2 \cite{tnl2k} & CVPR21
& 51.0 & - & 55.0 
& - & - & - 
& - & - & - 
& 42.0 & - & 42.0 
& 68.0 & - & 88.0 
& - & - & - 
\\

& SNLT \cite{snlt} & CVPR21
& 54.0 & 63.6 & 57.4 
& 26.2 & - & 30.0 
& - & - & - 
& 27.6 & - & 41.9 
& 66.6 & - & 80.4 
& 43.3 & 50.6 & 22.1 
\\

& Li et al. \cite{cmtr} & CVPRW22
& 53.0 & - & 56.0 
& - & - & - 
& - & - & - 
& 44.0 & 52.0 & 45.0 
& 69.0 & - & 91.0 
& - & - & - 
\\

& VLT$_{TT}$ \cite{vlt} & NeurIPS22
& 67.3 & - & 72.1 
& 48.4 & - & 55.9 
& 82.2 & - & 80.7 
& 53.1 & - & 53.3 
& 76.4 & - & 93.1 
& 69.4 & 81.1 & 64.5 
\\

& CiteTracker \cite{citetracker} & ICCV23
& 69.7 & 78.6 & 75.7 
& 45.9 & 59.7 & 54.0 
& \textcolor{blue}{\textbf{84.5}} & 89.0 & 84.2 
& 55.7 & 74.5 & 59.6 
& 69.6 & \textcolor{blue}{\textbf{85.1}} & 92.2 
& 74.7 & 84.3 & 73.0 
\\

& JoinNLT \cite{jointnlt} & CVPR23
& 60.4 & - & 63.6 
& 27.4 & 34.1 & 25.5 
& 66.7 & 69.2 & 59.4 
& 56.9 & - & 58.1 
& 65.3 & - & 85.6 
& 57.4 & 64.9 & 47.3 
\\

& VLT$_{OST-384}$ \cite{divert} & TPAMI24 
& 72.4 & - & 79.2 
& \textcolor{red}{\textbf{53.5}} & - & \textcolor{red}{\textbf{60.9}} 
& \textcolor{red}{\textbf{85.1}} & - & 83.7 
& 57.5 & - & 58.6 
& \textcolor{red}{\textbf{78.2}} & - & \textcolor{red}{\textbf{95.4}} 
& 76.3 & \textcolor{blue}{\textbf{87.1}} & 72.3 
\\

& UVLTrack-B \cite{UVLTrack} & CVPR24
& 69.4 & - & 74.9 
& 49.2 & - & 55.8 
& 83.4 & - & 82.1 
& \textcolor{red}{\textbf{62.7}} & - & \textcolor{red}{\textbf{65.4}} 
& 69.3 & - & 89.9 
& - & - & - 
\\

& QueryNLT \cite{context_aware} & CVPR24
& 59.9 & 69.6 & 63.5 
& - & - & - 
& - & - & - 
& 57.8 & \textcolor{blue}{\textbf{75.6}} & 58.7 
& 66.7 & 82.4 & 88.2 
& - & - & - 
\\

& MMTrack \cite{mmtrack} & CVPR24
& 70.0 & 82.3 & 75.7 
& 49.4 & 59.9 & 55.3 
& - & - & - 
& 58.6 & 75.2 & 59.4 
& 70.5 & - & 91.8 
& - & - & - 
\\

\specialrule{2pt}{0pt}{0pt}

& CLDTracker & Ours
& \textcolor{red}{\textbf{74.0}} & \textcolor{red}{\textbf{83.9}} & \textcolor{red}{\textbf{81.1}} 
& \textcolor{blue}{\textbf{53.1}} & \textcolor{red}{\textbf{64.8}} & \textcolor{blue}{\textbf{60.6}} 
& \textcolor{red}{\textbf{85.1}} & \textcolor{red}{\textbf{89.7}} & \textcolor{red}{\textbf{84.9}} 
& \textcolor{blue}{\textbf{61.5}} & \textcolor{red}{\textbf{82.2}} & \textcolor{blue}{\textbf{64.3}} 
& \textcolor{blue}{\textbf{77.8}} & \textcolor{red}{\textbf{86.1}} & \textcolor{blue}{\textbf{94.8}} 
& \textcolor{red}{\textbf{77.5}} & 85.4 & \textcolor{red}{\textbf{75.6}} 
\\

\specialrule{2pt}{0pt}{0pt}

\end{tabular}
}
\label{tab:res}
\end{table*}

\subsection{Quantitative Evaluation}
\noindent Table~\ref{tab:res} presents a comprehensive comparison between CLDTracker and SOTA methods across six prominent tracking benchmarks: LaSOT, LaSOT\textsubscript{ext}, TrackingNet, TNL2K, OTB99-Lang, and GOT-10k.
CLDTracker consistently demonstrates top-tier performance, ranking among the leading trackers across all datasets and metrics. While it may not dominate every individual metric, it achieves a balanced and robust performance profile. Specifically, excluding GOT-10k, CLDTracker ranks first in S and P on 2 out of 5 benchmarks and second on the remaining 3. Moreover, it secures the top rank in NP across all five benchmarks. On GOT-10k, CLDTracker achieves the highest scores in both AO and SR\textsubscript{0.75}, further underscoring its generalizability and competitiveness.

\noindent \textbf{(a) LaSOT}. CLDTracker achieves the best performance on the LaSOT benchmark across all three evaluation metrics, surpassing the previous SOTA tracker HIPTrack by 1.3\% in S, 1.0\% in NP, and 1.6\% in P.

\noindent \textbf{(b) LaSOT\textsubscript{ext}}. CLDTracker achieves highly competitive results on LaSOT\textsubscript{ext}, with S and P scores of 53.1\% and 60.6\%, respectively, just 0.4\% and 0.3\% behind the top-performing method VLT\textsubscript{OST-384}. Additionally, it attains the highest NP score of 64.8\%, outperforming the second-best method by 0.5\%.

\noindent \textbf{(c) TrackingNet}. CLDTracker delivers the strongest overall performance on the TrackingNet benchmark, achieving an S score of 85.1\%, NP of 89.7\%, and P of 84.9\%. It ties with VLT\textsubscript{OST-384} and ARTrack\textsubscript{384} in S while outperforming all other competitors by 0.6\% in NP and 0.1\% in P.

\noindent \textbf{(d) TNL2k}. On the TNL2K benchmark, CLDTracker achieves competitive scores of 61.5\% in S and 64.3\% in P, within 1.2\% and 1.1\% of the top-performing tracker VLT\textsubscript{OST-384}. Remarkably, it records the highest NP of 82.2\%, outperforming the next-best method by a substantial 6.6\%.

\noindent \textbf{(e) OTB99-Lang}. CLDTracker performs strongly on OTB99-Lang, achieving an S score of 77.8\% and a P score of 94.8\%, trailing the top method VLT\textsubscript{OST-384} by only 0.4\% and 0.6\%, respectively. It achieves the highest NP of 86.1\%, leading its closest competitor by 1.0\%.

\noindent \textbf{(f) GOT-10k.} On the GOT-10k benchmark, CLDTracker achieves an AO of 77.5\% and SR\textsubscript{0.75} of 75.6\%, surpassing HIPTrack by 0.13\% and 1.1\%, respectively. It also records a competitive SR\textsubscript{0.50} of 85.4\%, trailing the best-performing method by only 2.6\%.

\noindent \textbf{(g) VastTrack.} As shown in Table~\ref{tab:vasttrack_attribute_auc}, CLDTracker achieves SOTA performance across all attributes and on the overall test set of VastTrack. Notably, it outperforms the second-best model, CiteTracker, by 2.2\%, 2.3\%, and 2.4\% on the S, NP, and P metrics, respectively. These consistent gains highlight the strong generalization ability of CLDTracker to novel object classes and diverse tracking scenarios.

\begin{figure*}
  \centering
  \begin{subfigure}[b]{0.48\textwidth}
    \includegraphics[width=\textwidth, height=9cm]{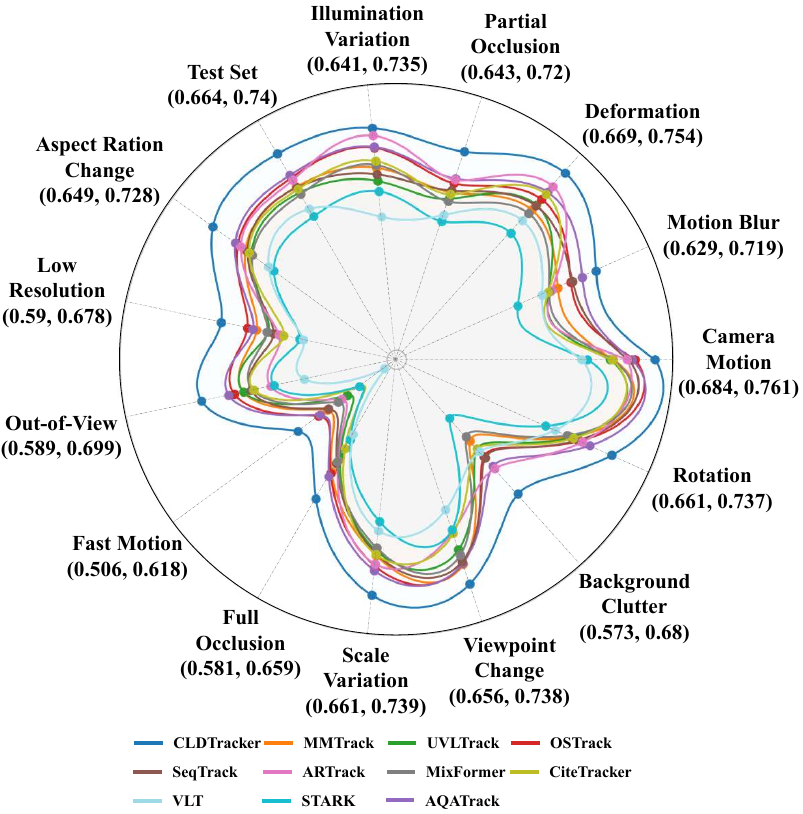}
    \caption{}
  \end{subfigure}
  \begin{subfigure}[b]{0.48\textwidth}
    \includegraphics[width=\textwidth, height=8.85cm]{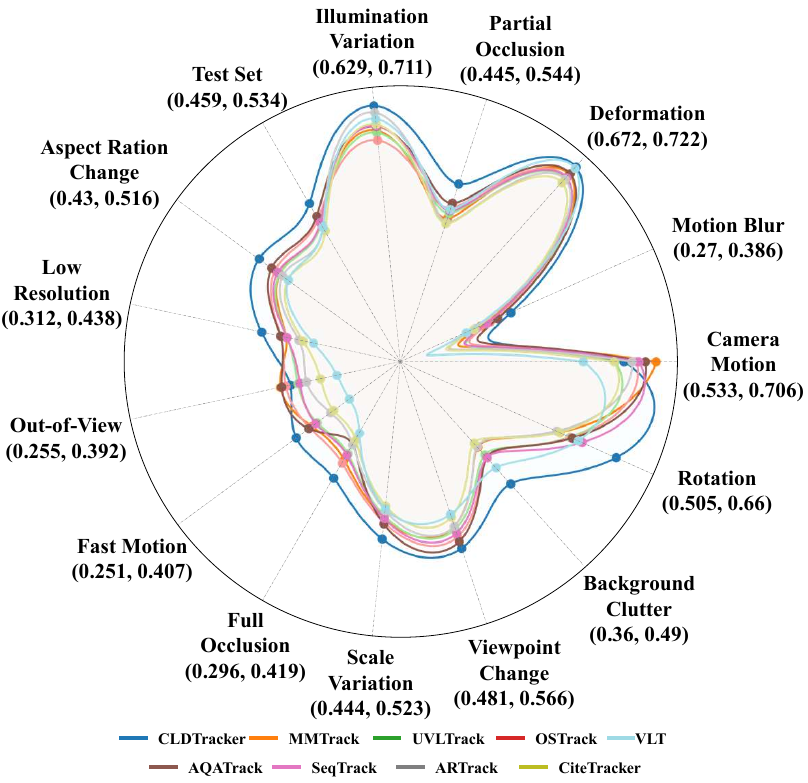}
    \caption{}
  \end{subfigure}

\caption{AUC scores of different attributes on (a) LaSOT, and (b) LaSOT\textsubscript{ext}. Best viewed in color.}

\label{fig:attributes}
\end{figure*}

\noindent \textbf{(h) Attribute-Wise Performance Analysis.} Figs. \ref{fig:attributes}(a) and (b) illustrate an attribute-wise evaluation on LaSOT and LaSOT\textsubscript{ext}, comparing CLDTracker with representative SOTA algorithms. The results demonstrate that CLDTracker outperforms competing trackers across most attributes, highlighting the effectiveness of incorporating high-level linguistic cues in the comprehensive language description.

\noindent Table \ref{tab:attribute_auc} presents a comprehensive evaluation of CLDTracker against the baseline VL tracker, CiteTracker \cite{citetracker}, across 14 tracking attributes on the LaSOT and LaSOT\textsubscript{ext} datasets. CLDTracker consistently outperforms the baseline across all attributes, underscoring its robustness and strong generalization capability. Specifically, our method achieves an average performance gain of 5.21\% on LaSOT and 8.74\% on LaSOT\textsubscript{ext}.

\begin{table*}[h]
\centering
\caption{Attribute-wise AUC (\%) comparison on LaSOT and LaSOT\textsubscript{ext} between CLDTracker and the baseline VL model \cite{citetracker}.}

\label{tab:attribute_auc}

\renewcommand{\arraystretch}{1.5}
\setlength{\tabcolsep}{5pt}

\resizebox{\linewidth}{!}{
\begin{tabular}{c|ccccccccccccccc}
    
\specialrule{2pt}{0pt}{0pt}
    
Method & ARC & BC & CM & DEF & FM & FOC & IV & LR & MB & OV & POC & ROT & VC & SV \\
    
\specialrule{2pt}{0pt}{0pt}
    
\begin{tabular}[l]{@{}c@{}}CiteTracker\\\cite{citetracker}\end{tabular} & 68.1 & 61.6 & 71.6 & 72.4 & 53.8 & 59.8 & 70.0 & 61.1 & 66.4 & 64.3 & 67.3 & 69.3 & 68.2 & 69.6 \\

CLDTracker & 72.8 & 68.0 & 76.1 & 75.4 & 61.8 & 65.9 & 73.5 & 67.8 & 71.9 & 69.9 & 72.0 & 73.7 & 73.8 & 73.9 \\

\rowcolor{gaincolor} \textbf{\% Gain} & +4.7\% & +6.4\% & +4.5\% & +3\% & +8\% & +6.1\% & +3.5\% & +6.7\% & +5.5\% & +5.6\% & +4.7\% & +4.4\% & +5.6\% & +4.3\% \\

\specialrule{1pt}{0pt}{0pt}

\begin{tabular}[l]{@{}c@{}}CiteTracker\\\cite{citetracker}\end{tabular} & 43.1 & 36.0 & 60.6 & 67.2 & 29.7 & 31.4 & 66.7 & 33.7 & 28.9 & 29.5 & 44.5 & 51.2 & 48.8 & 44.4 \\

CLDTracker & 51.6 & 49.0 & 63.0 & 72.2 & 40.7 & 41.9 & 71.1 & 43.8 & 38.6 & 36.8 & 54.4 & 66.0 & 56.6 & 52.3 \\

\rowcolor{gaincolor} \textbf{\% Gain} & +8.5\% & +13\% & +2.4\% & +5\% & +11\% & +10.5\% & +4.4\% & +10.1\% & +9.7\% & +7.3\% & +9.9\% & +14.8\% & +7.8\% & +7.9\% \\

\specialrule{2pt}{0pt}{0pt}
        
\end{tabular}
}
\end{table*}


We further evaluate CLDTracker on TNL2K \cite{tnl2k}, a challenging benchmark for language-guided tracking, using 17 attribute categories. As shown in Table \ref{tab:attribute_tnl2k_auc}, CLDTracker once again demonstrates consistent gains across all attributes, achieving an average improvement of 5.74\% over the baseline.
Notably, the TTFUM module contributes significantly to performance improvements on occlusion-related attributes such as Partial Occlusion (POC) and Full Occlusion (FOC). Moreover, CLDTracker benefits from its \(B_t\) and the Prompt Adapter, which boost performance on appearance-based challenges including Background Clutter (BC), Deformation (DEF), and Illumination Variation (IV). These findings highlight the effectiveness of many-to-one alignment and reinforce CLDTracker's adaptability and robustness in diverse and challenging tracking scenarios.

\begin{table*}[h]
\centering
\caption{Attribute-wise AUC (\%) comparison on TNL2K between CLDTracker and the baseline VL model \cite{citetracker}.}

\label{tab:attribute_tnl2k_auc}

\renewcommand{\arraystretch}{1.5}
\setlength{\tabcolsep}{5pt}

\resizebox{\linewidth}{!}{
\begin{tabular}{c|cccccccccccccccccc}

\specialrule{2pt}{0pt}{0pt}

Method & ARC & BC & CM & DEF & FM & FOC & IV & LR & MB & OV & POC & ROT & VC & SV & AS & TC & MS \\

\specialrule{2pt}{0pt}{0pt}

\begin{tabular}[l]{@{}c@{}}CiteTracker\\\cite{citetracker}\end{tabular} & 61.5 & 54.2 & 57.1 & 57.8 & 55.7 & 44.6 & 61.0 & 38.4 & 52.9 & 41.9 & 52.1 & 59.8 & 54.5 & 54.5 & 65.6 & 29.3 & 20.7 \\

CLDTracker & 64.9 & 59.7 & 62.3 & 62.6 & 61.6 & 50.6 & 66.8 & 49.2 & 58.3 & 47.6 & 58.2 & 64.0 & 59.7 & 59.8 & 68.6 & 39.6 & 30.9 \\

\rowcolor{gaincolor} \textbf{\% Gain} & +3.4\% & +5.5\% & +5.2\% & +4.8\% & +5.9\% & +6\% & +5.8\% & +10.8\% & +5.4\% & +5.7\% & +6.1\% & +4.2\% & +5.2\% & +5.3\% & +3\% & +10.3\% & +10.2\% \\

\specialrule{2pt}{0pt}{0pt}
        
\end{tabular}
}
\end{table*}


In particular, we emphasize that several of the evaluated attributes directly correspond to typical forms of visual noise including BC, MB, IV, POC, FOC, and Rotation (ROT). These attributes simulate real-world degradation scenarios such as cluttered scenes, lighting inconsistencies, motion-induced blur, and occlusions. As shown in Tables \ref{tab:attribute_auc} and \ref{tab:attribute_tnl2k_auc}, CLDTracker consistently surpasses the baseline under these noisy conditions, demonstrating its resilience to visual perturbations and further validating its robustness in unconstrained environments.

\begin{figure*}[h!]
    \centering    
    \includegraphics[width=0.99\textwidth]{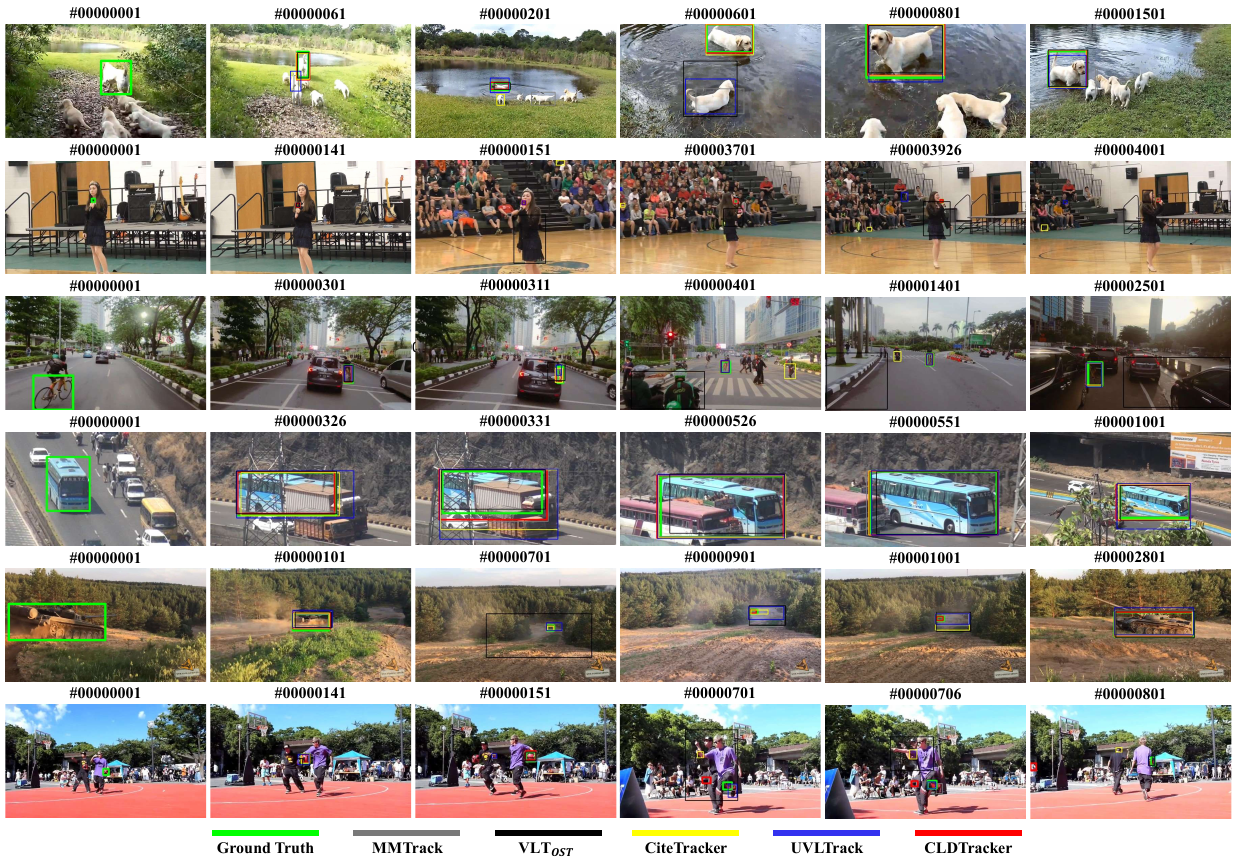}
    \caption{Results visualization of different trackers. The comparison shows that our CLDTracker could perform robust tracking under complex scenarios (e.g., Similar Interferences, Small Objects, Deformation, Occlusion, and  Disappearance), while the sixth row exhibits a failure case of our method when the target is occluded multiple times for a long period (with around 100 frames).}
 \label{fig:vr_failure}
 \end{figure*}

\subsection{Qualitative Evaluation}
\subsubsection{Visualization of Tracking Result and Failure Case}
\noindent As shown in Fig. \ref{fig:vr_failure}, the proposed CLDTracker demonstrates superior robustness in challenging scenarios, including interference from similar objects (first row), small objects amidst background clutter (second and fifth rows), and occlusion and deformation (third and fourth rows). It highlights the effectiveness of the learned multi-modal representation, especially in complex environments, where our tracker outperforms other SOTAs.

The sixth row shows the failure case of our tracker. In this case, the target is fully occluded for about 100 frames and distracted by similar objects, leading to ineffectiveness of our tracker in learning helpful information. A possible solution to deal with this is to apply a memory bank mechanism, and we leave this to future work.

\begin{figure*}[h!]
    \centering    
    \includegraphics[width=0.8\linewidth]{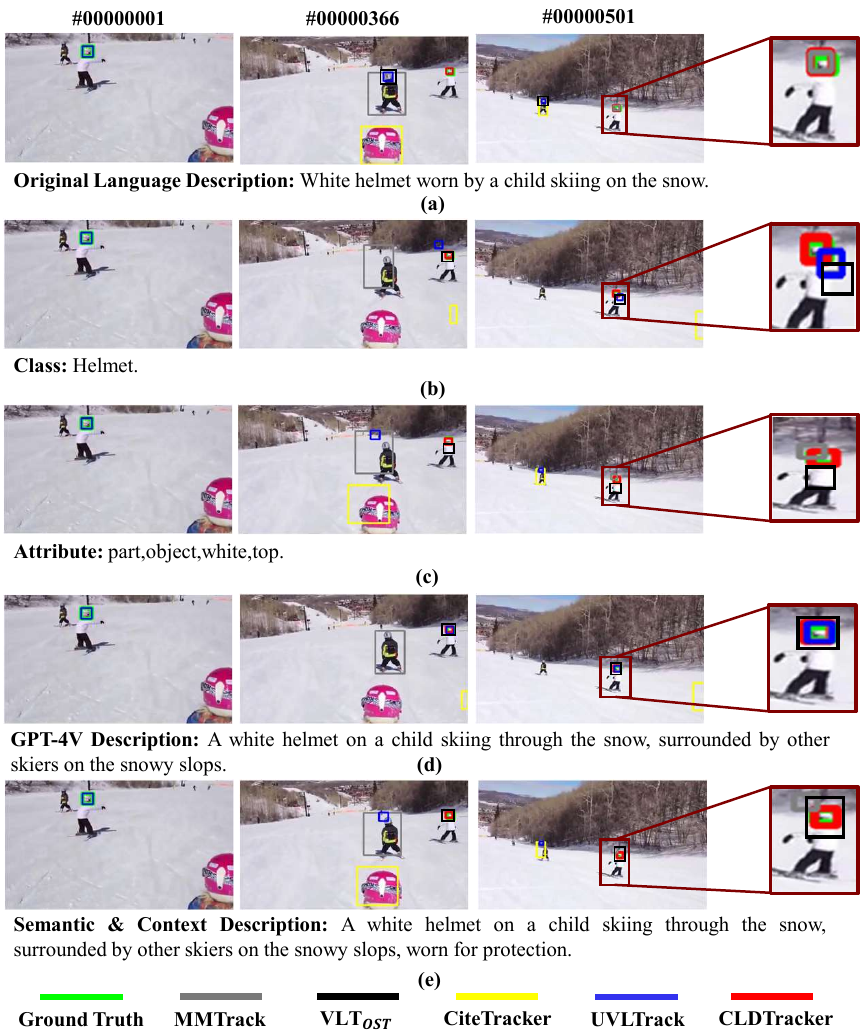}
    \caption{Visual results of various trackers under different input language descriptions. The comparison highlights the robustness of our CLDTracker, which mitigates sensitivity to variations in language annotation styles, maintaining focus on the overall semantic content rather than specific language nuances.}
 \label{fig:vr}
 \end{figure*}

Another visualization is shown in Fig. \ref{fig:vr}, where the proposed CLDTracker exhibits robust tracking performance across various language descriptions, including the official language description (first row), class names (second row), attribute descriptions (third row), GPT-4V descriptions (fourth row), and \(S\&C\) descriptions (fifth row). This demonstrates the effectiveness of its learned multi-modal representation, particularly in adapting to diverse and dynamic language cues.

\begin{figure*}[h!]
    \centering    
    \includegraphics[width=0.85\textwidth]{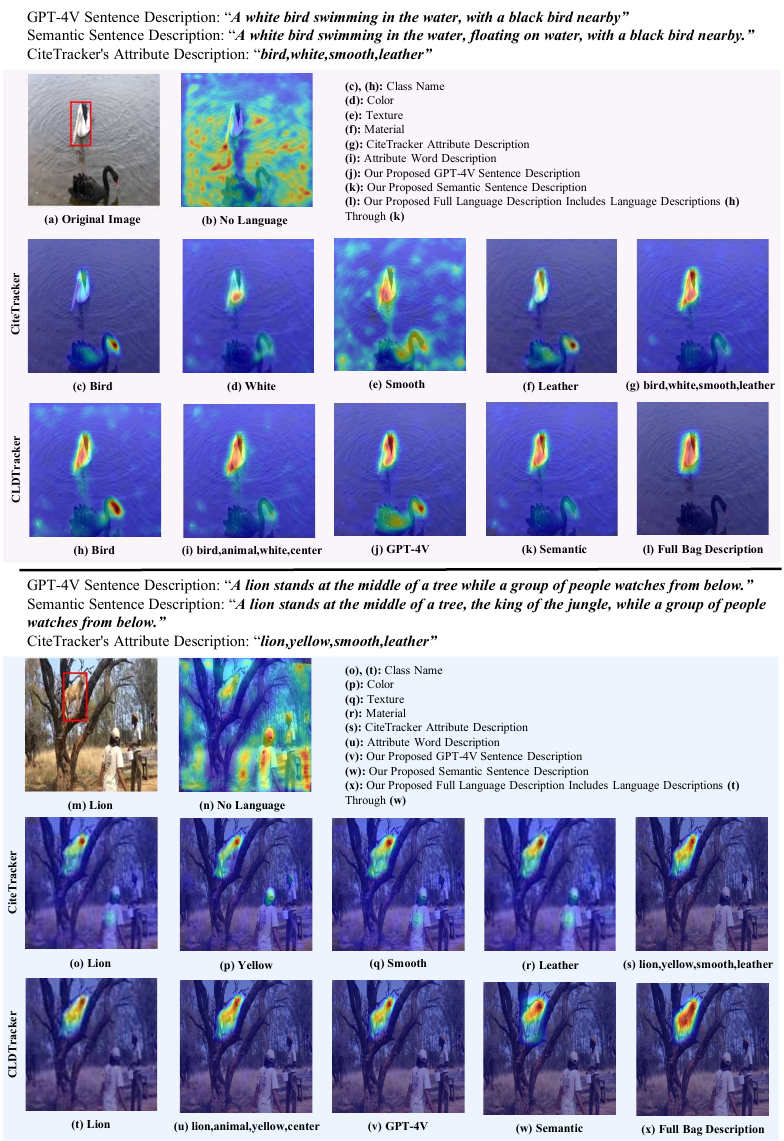}
    \caption{Activation visualizations of CLDTracker and CiteTracker using GradCAM \cite{gradcam} on two examples: a bird and a lion, each paired with different language descriptions. CLDTracker demonstrates clear target distinction from interfering elements across varied language descriptions.}
 \label{fig:heatmaps}
 \end{figure*}

\subsubsection{Activation Analysis of Different Language Descriptions} 
\noindent Language descriptions provide high-level semantic cues that enhance target-specific channels while suppressing irrelevant ones. As shown in Fig. \ref{fig:heatmaps}, we analyze the impact of different language inputs on our proposed CLDTracker compared to the baseline VL model, CiteTracker, demonstrating how language descriptions improve target discernibility.

The first and fourth rows depict the input image and CiteTracker's activation map with no language input, respectively. The second and fifth rows show activation maps generated by CiteTracker using their proposed attribute description annotations. The third and sixth rows illustrate the activation maps produced by CLDTracker, leveraging components of the \(B_t\) and the full \(B_t\). Notably, the baseline VL model frequently gets misled by objects of the same class. In contrast, CLDTracker consistently focuses on the object of interest (red-highlighted areas). The introduction of detailed attribute annotations, such as Figs. \ref{fig:heatmaps}(k) and (u), and \(S\&C\) descriptions, as shown in Figs. \ref{fig:heatmaps}(k) and (w), effectively eliminates distractor responses, showcasing the benefits of precise annotations for improved target localization.

Additionally, Fig. \ref{fig:heatmaps}(j) highlights that GPT-4V descriptions can occasionally be misled by objects of the same class. This occurs when the descriptions introduce irrelevant or extraneous details, resulting in misaligned visual and textual representations for tracking.

The final column displays activation maps generated using complete descriptions. It is evident that CLDTracker maintains a sharper focus on the target object compared to the baseline model, demonstrating its superior ability to accurately locate and track the target. These findings validate the effectiveness of our unified and adaptive VL representation and underscore the importance of detailed language descriptions for robust and precise target tracking.

\subsection{Ablation Studies}
\noindent To evaluate the contribution of each component in our tracker, we conduct ablation studies using various variants of CLDTracker, each designed to isolate and test the impact of individual components on the overall tracking performance.

\noindent \textbf{Baseline Model:} This model utilizes only the backbone to extract joint visual features from the target and test images. A prediction head, operates directly on the joint visual feature maps, is then applied to these features to estimate the final tracking result.

\noindent \textbf{Baseline VL model:} This variant introduces an image-text conversion model to generate category and attribute descriptions for both the template and search frames. These descriptions are correlated with the visual features to produce the features used for target state estimation.

\noindent \textbf{With Class (\(C\)):} This model variant focuses on predefined categories relevant to the target from the \(B_t\).

\noindent \textbf{With Attribute (\(A\)):} This variant considers predefined attributes \cite{divert} associated with the target from the \(B_t\).

\noindent \textbf{With GPT-4V Descriptions (\(GPT-4V\)):} This model only incorporates detailed descriptions of the target generated by GPT-4V from the \(B_t\).

\noindent \textbf{With Descriptions with Semantic and Context Information (\(S\&C\)):} This model uses a \(S\&C\) sequence which integrates both visual and semantic insights relevant to the target, extracted from the \(B_t\).

\noindent \textbf{With (\(C\)) and (\(A\)):} This variant combines both predefined categories and attributes related to the target from the \(B_t\).

\noindent \textbf{With (\(C\)), (\(A\)), and (\(GPT-4V\)):} In this variant, the predefined categories, attributes, and GPT-4V detailed descriptions are considered together for target representation from the \(B_t\).

\noindent \textbf{With the full \(B_t\) which includes (\(C\)), (\(A\)), (\(GPT-4V\)), and (\(S\&C\)):} This model variant uses the \(B_t\), including all predefined categories, attributes, GPT-4V descriptions, and \(S\&C\) descriptions.

\noindent \textbf{CLDTracker:} Our full model incorporates temporal variation by generating multiple textual descriptions for the search frames which allows it to adapt to changes over time. The Prompt Learner module allows CLDTracker to extract detailed image-conditioned textual descriptions for both the template and search frames. The temporal information is then added in the TTFUM by averaging \(t_o\) descriptions. These descriptions are then correlated with the visual features to form a rich, context-aware feature representation, which enhances the accuracy of target state estimation across frames. This ensures continuous update and alignment of textual and visual representations as the target's appearance evolves.

Note that not all model variants are evaluated in every experiment. Instead, each variant is selectively employed based on the specific component being analyzed, allowing for a focused evaluation of individual contributions within the overall framework.

The following modifications are applied to the baseline VL model to evaluate different components, starting with Prompt Adapter, which enhances VL alignment by dynamically selecting the most relevant textual description from the \(B_t\) using visual context, ensuring that the chosen prompts are tailored to the current image.

\begin{table}[h]
\centering
\caption{Impact of Window Size (WS) variation in the TTFUM module of the baseline VL model, both with (w/) and without (w/o) the Prompt Adapter. A WS of 1 corresponds to the absence of TTFUM. Performance is evaluated on four benchmarks: OTB99-Lang, LaSOT, LaSOT\textsubscript{ext}, and TNL2K. The top two results for each dataset are highlighted in \textcolor{red}{red} and \textcolor{blue}{blue}, respectively.}

\label{tab:ablation2}

\resizebox{0.99\linewidth}{!}{

\begin{tabular}{l|c|c|cc|cc|cc|cc|cc|cc}

\specialrule{2pt}{0pt}{0pt}

\multirow{2}{3em}{Dataset} & \multirow{2}{5em}{Evaluation Metric} & \multirow{2}{3em}{Baseline Model} & \multicolumn{12}{c}{Baseline VL Model (\(WS\))} \\

\cline{4-15}

& & & 1 (w/o) & 1 (w) & 2 (w/o) & 2 (w) & 5 (w/o) & 5 (w) & 10 (w/o) & 10 (w) & 15 (w/o) & 15 (w) & 20 (w/o) & 20 (w) \\

\specialrule{2pt}{0pt}{0pt}

& S & 67.5 & 65.5
& 69.6 & 66.3 & \textcolor{blue}{\textbf{69.9}} & 67.8 & \textcolor{red}{\textbf{71.0}} & 66.2 & 69.7 & 65.5 & 68.1 & 64.6 & 67.2 \\
OTB99- & NP & 82.7 & 81.1
& 85.1 & 81.4 & \textcolor{blue}{\textbf{85.2}} & 81.6 & \textcolor{red}{\textbf{85.3}} & 81.2 & 84.8 & 80.9 & 83.1 & 80.6  & 82.6 \\
Lang & P & 88.8 & 85.7
& 92.2 & 86.1 & \textcolor{blue}{\textbf{92.4}} & 86.5 & \textcolor{red}{\textbf{92.8}} & 85.9 & 92.0 & 85.6 & 91.5 & 85.2  & 89.2 \\

\specialrule{1pt}{0pt}{0pt}

& S & \textcolor{red}{\textbf{71.1}} & 67.2
& 69.7 & 67.6 & \textcolor{blue}{\textbf{70.8}} & 67.6 & \textcolor{red}{\textbf{71.1}} & 67.3 & 70.7 & 66.7 & 70.4 & 66.2  & 70.1 \\
LaSOT & NP & \textcolor{red}{\textbf{81.1}} & 76.1 
& 78.6 & 76.5 & 79.9 & 76.6 & \textcolor{blue}{\textbf{80.4}} & 76.4 & 79.7 & 76.2 & 78.5 & 76.1  & 78.1 \\
& P & \textcolor{blue}{\textbf{77.6}} & 74.3
& 75.7 & 74.6 & 77.1 & 74.8 & \textcolor{red}{\textbf{78.2}} & 74.5 & 77.4 & 74.3 & 76.8 & 74.1  & 76.3 \\

\specialrule{1pt}{0pt}{0pt}

& S & \textcolor{red}{\textbf{50.5}} & 42.4
& 45.9 & 42.8 & 46.4 & 43.0 & \textcolor{blue}{\textbf{47.3}} & 42.7 & 46.8 & 42.4 & 46.2 & 42.4 & 45.7
\\
LaSOT\textsubscript{ext} & NP & \textcolor{red}{\textbf{61.3}} & 56.7
& 59.7 & 57.0 & 59.8 & 57.1 & \textcolor{blue}{\textbf{60.5}} & 56.8 & 59.3 & 56.4 & 58.9 & 56.3 & 58.7 
\\
& P & \textcolor{red}{\textbf{57.6}} & 52.0
& 54.0 & 52.3 & 54.5 & 52.6 & \textcolor{blue}{\textbf{54.9}} & 52.1 & 54.3 & 52.1 & 54.0 & 51.9 & 53.6
\\

\specialrule{1pt}{0pt}{0pt}

& S & 55.9 & 53.1
& 57.7 & 53.4 & \textcolor{blue}{\textbf{57.9}} & 53.7 & \textcolor{red}{\textbf{58.0}} & 53.2 & 57.3 & 53.1 & 57.1 & 52.9 & 56.8 
\\
TNL2K & NP & - & 70.5
& 74.5 & 70.9 & \textcolor{blue}{\textbf{74.7}} & 70.9 & \textcolor{red}{\textbf{75.0}} & 70.7 & \textcolor{blue}{\textbf{74.7}} & 70.5 & 74.1 & 70.3 & 73.8 
\\
& P & 57.1 & 55.8
& 59.6 & 56.2 & \textcolor{blue}{\textbf{59.7}} & 56.4 & \textcolor{red}{\textbf{59.9}} & 56.0 & 58.8 & 55.9 & 58.3 & 55.7 & 57.9
\\

\specialrule{2pt}{0pt}{0pt}

\end{tabular}
}
\end{table}

\subsubsection{Prompt Adapter}
\noindent \textbf{(a) Prompt Adapter Contribution:}
\noindent The Prompt Adapter enhances VL alignment by dynamically refining and selecting the most relevant attribute-based textual descriptions generated by the baseline VL model.
In Table~\ref{tab:ablation2}, columns labeled (w/) include the Prompt Adapter, while those marked (w/o) omit it. The Window Size (\(WS\)) controls the number of search images used to model temporal variations, with \(WS=1\) corresponding to the baseline VL model where TTFUM is disabled and only a single search region is analyzed.
In the absence of the Prompt Adapter (i.e., the (w/o) columns), similarity is computed directly between the attribute embeddings produced by the baseline VL model and the visual features extracted from the template and search images.
To isolate the effect of the Prompt Adapter, we evaluate it in the base VL setting, excluding both the \(B_t\) and TTFUM (\(WS=1\)). From Table \ref{tab:ablation2} we observe an average increase of 3.675\% in S, 3.375\% in NP, and 3.425\% in P, respectively, upon introducing the Prompt Adapter, demonstrating its effectiveness in improving VL alignment.

\noindent \textbf{(b) Evaluating Prompt Adapter Performance Under Temporal Modeling with TTFUM:}
Next, we enable the TTFUM by setting \(WS>1\), and evaluate the contribution of the Prompt Adapter by comparing model performance with (w/) and without (w/o) its inclusion.
As shown in Table~\ref{tab:ablation2}, the addition of the Prompt Adapter consistently enhances performance across all window sizes. Specifically, it achieves average gains of 3.66\% in S, 3.65\% in NP, and 3.35\% in P, demonstrating its effectiveness when integrated with temporal modeling.

\noindent \textbf{(c) Generalization of the Prompt Adapter Across Variants from the Bag of Textual Descriptions:}
In this experiment, we assess the impact of the Prompt Adapter when the baseline VL model's attribute-based descriptions are replaced with different components from the \(B_t\). This setup allows us to evaluate how effectively the Prompt Adapter adapts to a variety of textual input types.

We compare performance with (w/) and without (w/o) the Prompt Adapter across multiple textual description variants. As shown in Table~\ref{tab:ablation1}, incorporating the Prompt Adapter consistently improves performance across all configurations, achieving average gains of 2.04\% in S, 2.39\% in NP, and 2.34\% in P. These results demonstrate the module’s strength in refining textual representations and enhancing VL alignment, regardless of the input description type.

\subsubsection{Temporal Text Feature Update Mechanism}
\noindent \textbf{(a) TTFUM Contribution:}
\noindent To handle changes across frames, this modification computes the temporal difference in feature embeddings by comparing the similarity between the template embedding and the temporal variation in the search image embedding. The number of search images considered for temporal change is determined by the \( WS \), where \(WS = 1\) corresponds to the baseline VL model, meaning only one search region is evaluated.
To isolate the contribution of the TTFUM, the following experiments remove the \(B_t\) and instead rely solely on the baseline VL model's category and attribute descriptions.

Table \ref{tab:ablation2} analyzes the influence of the \( WS \) in the TTFUM both with (w/) and without (w/o) the Prompt Adapter, which reduces the impact of noisy or inaccurate text descriptions from the current frame by using historical text features. Moderate \( WS \) values (e.g., \( WS = 2 \) or \( WS = 5 \)) consistently deliver the best results across datasets.
For instance, with the Prompt Adapter enabled, on the OTB99-Lang dataset, the S increases from 69.6 at \( WS = 1 \) to a peak of 71.0 at \( WS = 5 \), before declining at larger \( WS \) values. Similarly, on the LaSOT dataset, the S reaches its maximum of 71.1 at \( WS = 5 \).
On the LaSOT\(_{ext}\) and TNL2K datasets, \( WS = 5 \) also achieves the highest performance across all metrics.
This trend remains consistent even when the Prompt Adapter is removed, indicating that a moderate WS \( WS = 5 \) provides an optimal balance between temporal context and model stability.
However, as \( WS \) increases beyond a certain point, performance begins to decline due to the inclusion of outdated temporal information that may no longer align with the object’s evolving appearance. This over-reliance on long-term context introduces feature drift and weakens visual-textual alignment. These results highlight that moderate \( WS \) values strike a balance between temporal stability and adaptability to current-frame variations, avoiding degradation caused by excessive historical dependence.

\begin{figure*}[h!]
    \centering    
    \includegraphics[width=0.99\linewidth]{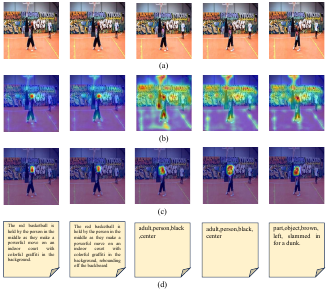}
    \caption{Visual results demonstrating the robustness of our proposed TTFUM against occlusion. The sequence progresses from left to right, where the leftmost frame represents the previous frame, and subsequent frames depict the next ones. (a) Input image, (b) Activation visualization without the proposed module, (c) Activation visualization with the proposed module, and (d) Selected input text by the Prompt Adapter module.}
 \label{fig:TTFU}
 \end{figure*}

Fig. \ref{fig:TTFU} illustrates the selected frames from a video sequence progressing from left to right, with the leftmost frame representing the previous frame and subsequent frames showing the following ones. Even when the Prompt Adapter module selects inaccurate text due to occlusion (Fig. \ref{fig:TTFU}(d)), the proposed CLDTracker, enhanced by the TTFUM (Fig. \ref{fig:TTFU}(c)), effectively leverages prior text knowledge, demonstrating superior performance compared to the version without the proposed module (Fig. \ref{fig:TTFU}(b)).

\begin{table}[h]
\centering
\begin{minipage}{0.48\textwidth}
\centering
\caption{Comparison of different temporal aggregation strategies.}
\label{tab:ttfum_aggregation}
\begin{tabular}{c|ccc}
\specialrule{2pt}{0pt}{0pt}
Aggregation Method & S & NP & P \\
\specialrule{2pt}{0pt}{0pt}
Last Frame & 69.7 & 78.6 & 75.7 \\
Max & 70.3 & 78.9 & 76.8 \\
Weighted Average & \textcolor{blue}{\textbf{70.8}} & \textcolor{blue}{\textbf{79.7}} & \textcolor{blue}{\textbf{77.5}} \\
Average (Ours) & \textcolor{red}{\textbf{71.1}} & \textcolor{red}{\textbf{80.4}} & \textcolor{red}{\textbf{78.2}} \\
\specialrule{2pt}{0pt}{0pt}
\end{tabular}
\end{minipage}
\hfill
\begin{minipage}{0.48\textwidth}
\centering
\caption{Comparison of different attention weight update intervals.}
\label{tab:update_intervals}
\begin{tabular}{c|ccc}
\specialrule{2pt}{0pt}{0pt}
Update Interval & S & NP & P \\
\specialrule{2pt}{0pt}{0pt}
Every 2 Frames & \textcolor{blue}{\textbf{70.5}} & \textcolor{blue}{\textbf{79.1}} & \textcolor{blue}{\textbf{77.0}} \\
Every 3 Frames & 69.8 & 78.4 & 76.2 \\
Every 4 Frames & 69.3 & 78.0 & 75.6 \\
Every 5 Frames & 68.9 & 77.0 & 75.1 \\
Every Frame (Ours) & \textcolor{red}{\textbf{71.1}} & \textcolor{red}{\textbf{80.4}} & \textcolor{red}{\textbf{78.2}} \\
\specialrule{2pt}{0pt}{0pt}
\end{tabular}
\end{minipage}
\end{table}

\noindent \textbf{(b) Impact of Aggregation Strategies in TTFUM:}
\noindent To evaluate the influence of different temporal aggregation strategies within TTFUM, we designed multiple variants that govern how search features from past frames are aggregated, where the default approach uses simple averaging, as defined in Eq.~\ref{eq:ttfum}. Specifically, we investigate three alternatives: (a) using only the most recent frame without any temporal averaging, corresponding to the baseline VL model, where \(W_{att} = \textrm{softmax}\Bigg(- \Big|\tilde{T}_{e} - \tilde{T}_{s}(t)\Big|\Bigg)\), (b) selecting the maximum response across the past frames, defined as \(W_{att} = \textrm{softmax}\Bigg(- \Big|\tilde{T}_{e} - \max_{i=t-n+1}^{t} \tilde{T}_{s}(i)\Big|\Bigg)\), and (c) applying a weighted sum with temporal decay to emphasize recent frames, formulated as \(W_{att} = \textrm{softmax}\Bigg(- \Big|\tilde{T}_{e} - \sum_{i=t-n+1}^{t} w_i \tilde{T}_{s}(i)\Big|\Bigg), \quad \sum w_i = 1,\), where \(w_i\) are exponentially decaying weights. These strategies modify the computation of attention weights \(W_{att}\) as defined in Eq. \ref{eq:ttfum}. All experiments use a temporal window size of \(WS=5\).

As summarized in Table\ref{tab:ttfum_aggregation}, the average aggregation strategy, where search features from the past \(n=5\) frames are equally weighted, yields the best overall tracking performance, achieving a S of 71.1\%, a NP of 80.4\%, and a P of 78.2\%. The weighted average variant performs comparably, slightly trailing the Average method. In contrast, the maximum operator, while providing robustness to abrupt appearance changes, demonstrates slightly less stable behavior, resulting in a minor drop in performance. The last frame only variant, which disregards temporal smoothing by relying solely on the most recent frame (\(n=1\)), achieves the lowest performance across all metrics. These findings highlight the importance of aggregating multiple frames to mitigate frame-level noise and enhance tracking robustness in dynamic environments.

\noindent \textbf{(c) Impact of Attention Weight Update Intervals in TTFUM:}
\noindent We study how the frequency of attention weight \(W_{attn}\) updates affects tracking performance. By default, \(W_{attn}\) are updated at every frame, as defined in Eq.\ref{eq:ttfum}. Using the average aggregation strategy with \(n=5\) frames fixed across all settings, we vary the update interval to control how often the attention weights \(W_{attn}\) are recalculated in response to incoming frames.
As presented in Table~\ref{tab:update_intervals}, updating the attention weights at every frame achieves the highest tracking performance, reaching a S of 71.1\%, a NP of 80.4\%, and a P of 78.2\%. Less frequent updates, such as every 2, 3, 4, or 5 frames, slightly enhance noise robustness but compromise the model’s responsiveness to appearance changes, resulting in gradual performance degradation. 
These results suggest that recalculating \(W_{attn}\) at every frame is critical for maintaining a balance between adaptability and stability, and we adopt this setting in the final model configuration.

\begin{table*}[ht]
\centering
\caption{Effect of different language inputs in the baseline VL model, both with (w/) and without (w/o) the Prompt Adapter, and the stepwise development of the \(B_t\) across four datasets: OTB99-Lang, LaSOT, LaSOT\textsubscript{ext}, and TNL2K, both with (w) and without (w/o) the Prompt Adapter. \(C\) represents the class name, \(A\) indicates the attributes, and \(S\&C\) refers to the descriptions with \(S\&C\). The best two results are shown in \textcolor{red}{red}, and \textcolor{blue}{blue} color, respectively (excluding CLDTracker column).}

\label{tab:ablation1}

\resizebox{0.99\textwidth}{!}{

\begin{tabular}{c|c|c|c|cc|cc|cc|cc|cc|cc|cc|c}

\specialrule{2pt}{0pt}{0pt}

\multirow{2}{3em}{Dataset} & \multirow{2}{5em}{\begin{tabular}[l]{@{}c@{}}Evaluation\\Metric\end{tabular}} & \multirow{2}{3em}{\begin{tabular}[c]{@{}c@{}}Base\\Model\end{tabular}} & \multirow{2}{4em}{\begin{tabular}[c]{@{}c@{}}Base VL\\Model\end{tabular}} & \multicolumn{2}{c|}{\(C\)} & \multicolumn{2}{c|}{\(A\)} & \multicolumn{2}{c|}{\(GPT-4V\)} & \multicolumn{2}{c|}{\(S\&C\)} & \multicolumn{2}{c|}{\begin{tabular}[c]{@{}c@{}} \(C\) \& \(A\) \end{tabular}} & \multicolumn{2}{c|}{\begin{tabular}[c]{@{}c@{}}\(C\) \& \(A\) \\ \& \(GPT-4V\)\end{tabular}} & \multicolumn{2}{c|}{\begin{tabular}[c]{@{}c@{}}\(C\) \& \(A\) \& \\ \(GPT-4V\) \& \(S\&C\)\end{tabular}} & \multirow{2}{5em}{CLDTracker} \\

\cline{5-18}

& & & & w/o & w & w/o & w & w/o & w & w/o & w & w/o & w & w/o & w & w/o & w \\

\specialrule{2pt}{0pt}{0pt}

& S & 67.5 & 69.6 
& 67.7 & 69.7 & 68.0 & 70.0 & 68.2 & 70.2 & 70.4 & 72.4 & 69.6 & 71.6 & 71.8 & \textcolor{blue}{\textbf{73.8}} & 73.9 & \textcolor{red}{\textbf{75.9}} & 77.8 \\
OTB99- & NP & 82.7 & 85.1 
& 83.0 & 85.5 & 83.2 & 85.7 & 83.3 & \textcolor{blue}{\textbf{85.8}} & 83.4 & \textcolor{red}{\textbf{85.9}} & 83.4 & \textcolor{red}{\textbf{85.9}} & 83.4 & \textcolor{red}{\textbf{85.9}} & 83.4 & \textcolor{red}{\textbf{85.9}} & 86.1 \\
Lang & P & 88.8 & 92.2 
& 90.0 & 92.4 & 90.4 & 92.9 & 90.6 & 93.1 & 91.0 & 93.5 & 90.9 & 93.4 & 91.1 & \textcolor{blue}{\textbf{93.6}} & 91.3 & \textcolor{red}{\textbf{93.8}} & 94.8 \\

\specialrule{1pt}{0pt}{0pt}

& S & 71.1 & 69.7 
& 67.5 & 70.1 & 67.7 & 70.3 & 68.1 & 70.7 & 68.8 & \textcolor{blue}{\textbf{71.4}} & 67.7 & 70.3 & 68.3 & 70.9 & 69.9 & \textcolor{red}{\textbf{72.5}} & 74.0 \\
LaSOT & NP & \textcolor{red}{\textbf{81.1}} & 78.6 
& 76.5 & 78.7 & 76.7 & 78.9 & 77.7 & 79.7 & 78.2 & 80.2 & 77.5 & 79.5 & 77.9 & 79.9 & 78.5 & \textcolor{blue}{\textbf{80.5}} & 81.1 \\
& P & 77.6 & 75.7 
& 73.7 & 75.7 & 73.9 & 75.9 & 75.4 & 77.4 & 76.2 & \textcolor{blue}{\textbf{78.2}} & 74.7 & 76.7 & 75.0 & 77.7 & 78.8 & \textcolor{red}{\textbf{80.8}} & 83.9 \\

\specialrule{1pt}{0pt}{0pt}

& S & \textcolor{red}{\textbf{50.5}} & 45.9 
& 42.3 & 44.3 & 42.9 & 44.9 & 43.3 & 45.3 & 45.7 & 47.7 & 44.1 & 46.1 & 46.5 & 48.5 & 48.4 & \textcolor{blue}{\textbf{50.4}} & 53.1 \\
LaSOT\textsubscript{ext} & NP & \textcolor{blue}{\textbf{61.3}} & 59.7 
& 55.4 & 57.4 & 56.2 & 58.2 & 56.4 & 58.4 & 57.9 & 59.9 & 57.2 & 59.2 & 57.2 & 59.9 & 58.9 & \textcolor{red}{\textbf{61.6}} & 64.8 \\
& P & \textcolor{blue}{\textbf{57.6}} & 54.0 
& 50.7 & 52.7 & 50.8 & 52.8 & 50.8 & 52.8 & 52.7 & 54.7 & 51.8 & 53.8 & 52.0 & 54.0 & 56.4 & \textcolor{red}{\textbf{58.4}} & 60.6 \\

\specialrule{1pt}{0pt}{0pt}

& S & 55.9 & 57.7 
& 55.6 & 57.6 & 55.4 & 57.4 & 55.9 & 57.9 & 55.9 & 57.9 & 57.1 & 58.1 & 57.3 & \textcolor{blue}{\textbf{58.3}} & 58.7 & \textcolor{red}{\textbf{59.7}} & 61.5 \\
TNL2K & NP & - & 74.5 
& 71.5 & 74.2 & 71.5 & 74.2 & 72.1 & 74.8 & 72.3 & 75.0 & 73.1 & 75.8 & 75.3 & \textcolor{blue}{\textbf{78.3}} & 77.5 & \textcolor{red}{\textbf{80.5}} & 82.2 \\
& P & 57.1 & 59.6 
& 56.1 & 58.8 & 56.6 & 59.3 & 57.4 & 60.1 & 57.7 & 60.4 & 57.5 & 60.2 & 57.7 & \textcolor{blue}{\textbf{60.6}} & 58.5 & \textcolor{red}{\textbf{61.5}} & 64.3 \\

\specialrule{2pt}{0pt}{0pt}

\end{tabular}
}
\end{table*}

\subsubsection{Comprehensive Bag of Textual Descriptions}
\noindent \textbf{(a) Individual and Incremental Impact of Text Components on Tracker Performance:} 
\noindent The following ablations assess both the individual contribution and the incremental benefits of components within \(B_t\) which demonstrates how each element enhances tracking performance. The analysis examines classes (\(C\)), attributes (\(A\)), GPT-4V descriptions, and descriptions with \(S\&C\) to highlight their individual value while also evaluating the cumulative impact of progressively incorporating these components into a \(B_t\). The results are presented in Table \ref{tab:ablation1}. For clarity, we have categorized the findings below.

\paragraph{(i) Impact of Vision and Text Feature Correlation:} The performance gap between the baseline model, and our CLDTracker highlights the significant advantage of correlating visual and text-based features for tracking tasks.

\noindent \paragraph{(ii) Impact of Using Single Text Description:} The performance of the baseline VL model improves significantly when leveraging class names \(C\), attributes \(A\), GPT-4V descriptions, and particularly \(S\&C\). Under the \(S\&C\) approach, the baseline VL model achieves an average 6.5\% increase in S across all four benchmarks compared to the baseline. This highlights the importance of incorporating richer detailed descriptions which help the model to better understand and track targets.

\noindent \paragraph{(iii) Incremental Impact of Text Components on Tracker Performance:} The results highlight the performance gains achieved by incrementally adding text components, \(C\), \(A\), GPT-4V descriptions, and \(S\&C\), to the \(B_t\). Each addition consistently improved performance across all benchmarks. 
For example, on the LaSOT dataset, the S increased from 69.7 for the baseline VL model to 72.5 with all components combined. Similarly, on OTB99-Lang, the S rose from 69.6 with the baseline VL model to 75.9 with the \(B_t\).

\noindent \paragraph{(iv) Component-Wise Contribution of the Semantic and Context Information (\(S\&C\)) Enrichment Module:} 
\noindent To assess the impact of each component in our \(S\&C\) enrichment pipeline, we conduct a stepwise evaluation by progressively incorporating its four key stages into the baseline VL model including synonym identification, semantic perturbation, task-specific descriptive phrases, and high-level conceptual abstraction. As shown in Table~\ref{tab:ablation_text}, each stage contributes incremental gains, with the complete integration of all components delivering the highest overall performance across datasets. This confirms the complementary nature of the individual enhancements in enriching the textual representation for visual tracking.

\begin{table}[h]
\centering
\caption{Incremental contribution of each stage in the \(S\&C\) enrichment module. Performance is reported in terms of S.}

\label{tab:ablation_text}
\begin{tabular}{c|cccc}
\specialrule{2pt}{0pt}{0pt}
\begin{tabular}[c]{@{}c@{}}Augmentation\\Strategy\end{tabular} & OTB99-Lang & LaSOT & LaSOT\(_{ext}\) & TNL2K \\
\specialrule{2pt}{0pt}{0pt}
Synonyms                          & 69.8 & 69.1 & 45.8 & 54.9 \\

\specialrule{1pt}{0pt}{0pt}

+ \begin{tabular}[c]{@{}c@{}}Semantic\\Perturbation\end{tabular}           & 70.4 & 69.7 & 46.3 & 55.6 \\

\specialrule{1pt}{0pt}{0pt}

+ \begin{tabular}[c]{@{}c@{}}Task-specific\\Phrases\end{tabular}  & \textcolor{blue}{\textbf{71.1}} & \textcolor{blue}{\textbf{70.4}} & \textcolor{blue}{\textbf{46.9}} & \textcolor{blue}{\textbf{56.7}} \\

\specialrule{1pt}{0pt}{0pt}

+ \begin{tabular}[c]{@{}c@{}}High-level\\Concept\end{tabular}              & \textcolor{red}{\textbf{72.4}} & \textcolor{red}{\textbf{71.4}} & \textcolor{red}{\textbf{47.7}} & \textcolor{red}{\textbf{57.9}} \\

\specialrule{2pt}{0pt}{0pt}
\end{tabular}
\end{table}

\noindent \textbf{(b) Comprehensive Bag of Textual Descriptions Quality:}
\noindent We conduct a series of ablation studies to evaluate the quality of the constructed \(B_t\), focusing on four aspects: the choice of generative model, the effect of textual granularity, the contribution of the validation pipeline, and the system’s robustness to incorrect or misleading language cues.

\noindent \paragraph{(i) Effect of Generative Model for Descriptions:}
\noindent To assess the impact of the generative model on producing target-specific (image-to-text), task-specific, and high-level contextual descriptions, we compare GPT-4V with two strong alternatives: BLIP-2 \cite{blip2} and the multi-modal LLaMA 3.2-90B model \cite{llama}, which accepts both text and image inputs and contains approximately 90 billion parameters.
For a fair comparison, a consistent filtering procedure, based on the steps outlined in Sec. 3.2.1(e), is applied to all models. 
As shown in Table \ref{tab:ablation_genmodels}, GPT-4V consistently outperforms the alternatives across all benchmarks. Its ability to generate diverse, semantically coherent, and contextually rich descriptions results in more informative textual embeddings, leading to significant improvements in tracking performance. These findings highlight the clear advantage of leveraging GPT-4V for textual enrichment in VL tracking tasks.

\begin{table}[h]
\centering

\caption{Comparison of generative models used for producing GPT-4V-generated descriptions and \(S\&C\). Results are reported in terms of S across four benchmarks: OTB99-Lang, LaSOT, LaSOT\textsubscript{ext}, and TNL2K.}

\label{tab:ablation_genmodels}
\begin{tabular}{c|cccc}
\specialrule{1pt}{0pt}{0pt}
Generative Model & OTB99-Lang & LaSOT & LaSOT\(_{ext}\) & TNL2K \\
\specialrule{1pt}{0pt}{0pt}
BLIP-2 \cite{blip2} & 66.7 & 66.2 & 42.0 & 54.1 \\
LLaMA 3.2 \cite{llama} & 67.4 & 67.0 & 42.6 & 55.0 \\
\specialrule{1pt}{0pt}{0pt}
GPT-4V \cite{gpt4v} (Ours) & \textcolor{red}{\textbf{68.2}} & \textcolor{red}{\textbf{68.1}} & \textcolor{red}{\textbf{43.3}} & \textcolor{red}{\textbf{55.9}} \\
\specialrule{1pt}{0pt}{0pt}
\end{tabular}
\end{table}

\noindent \paragraph{(ii) Impact of Textual Granularity:}
\noindent The level of detail in textual descriptions can significantly influence the effectiveness of VL tracking. Overly brief descriptions may lack sufficient semantic cues, while excessively verbose ones may introduce noise and redundancy. To better understand this trade-off, we conducted an ablation study varying the length and detail of GPT-4V-generated descriptions \cite{gpt4v}. Descriptions were grouped into three categories: concise (\(\leq\)10 tokens), moderately detailed (10-50 tokens), and verbose (\(>\)50 tokens).

As shown in Table~\ref{tab:ablation-granularity}, moderately detailed descriptions consistently outperformed both concise and verbose versions across multiple benchmarks. These findings confirm that while rich textual information supports more accurate tracking, excessive verbosity can dilute important signals and slightly degrade performance. Maintaining a balance between informativeness and brevity is therefore crucial when designing textual prompts for tracking.

\begin{table}[h]

\centering

\caption{Effect of GPT-4V-generated description granularity on tracking performance. Results are reported in terms of S, NP, and P across four benchmarks: OTB99-Lang, LaSOT, LaSOT\textsubscript{ext}, and TNL2K.}

\label{tab:ablation-granularity}
    

\begin{tabular}{c|c|ccc}

\specialrule{2pt}{0pt}{0pt}

\multirow{3}{3em}{Dataset} & \multirow{3}{5em}{Evaluation Metric} & \multicolumn{3}{c}{GPT-4V Description Length} \\

\cline{3-5}
    
& & \begin{tabular}[c]{@{}c@{}}Concise\\(tokens\(\leq\)10)\end{tabular} & \begin{tabular}[c]{@{}c@{}}Moderate\\(10\(<\)tokens\(\leq\)50)\end{tabular} & \begin{tabular}[c]{@{}c@{}}Verbose\\(tokens\(>\)50)\end{tabular} \\

\specialrule{2pt}{0pt}{0pt}

& S & \textcolor{blue}{\textbf{69.5}} & \textcolor{red}{\textbf{70.2}} & 68.6 \\
OTB99- & NP & \textcolor{blue}{\textbf{85.1}} & \textcolor{red}{\textbf{85.8}} & 84.7 \\
Lang & P & \textcolor{blue}{\textbf{92.2}} & \textcolor{red}{\textbf{93.1}} & 91.8 \\

\specialrule{1pt}{0pt}{0pt}

& S & \textcolor{blue}{\textbf{69.6}} & \textcolor{red}{\textbf{70.7}} & 69.1 \\
LaSOT & NP & \textcolor{blue}{\textbf{78.5}} & \textcolor{red}{\textbf{79.7}} & 78.2 \\
& P & \textcolor{blue}{\textbf{75.5}} & \textcolor{red}{\textbf{77.4}} & 75.2 \\

\specialrule{1pt}{0pt}{0pt}

& S & \textcolor{blue}{\textbf{44.5}} & \textcolor{red}{\textbf{45.3}} & 44.3 \\
LaSOT\textsubscript{ext} & NP & \textcolor{blue}{\textbf{57.7}} & \textcolor{red}{\textbf{58.4}} & 57.4 \\
& P & \textcolor{blue}{\textbf{52.7}} & \textcolor{red}{\textbf{52.8}} & 52.5 \\

\specialrule{1pt}{0pt}{0pt}

& S & \textcolor{blue}{\textbf{57.5}} & \textcolor{red}{\textbf{57.9}} & 57.3 \\
TNL2K & NP & \textcolor{blue}{\textbf{74.3}} & \textcolor{red}{\textbf{74.8}} & \textcolor{blue}{\textbf{74.3}} \\
& P & \textcolor{blue}{\textbf{59.0}} & \textcolor{red}{\textbf{60.1}} & 58.9 \\

\specialrule{2pt}{0pt}{0pt}

\end{tabular}
\end{table}

\noindent \paragraph{(iii) Impact of Validation Pipeline on the Bag of Textual Descriptions:}
\noindent To ensure the quality and relevance of the \(B_t\), particularly the GPT-4V-generated descriptions and synonyms, we employed a multi-stage control and validation pipeline addressing both content accuracy and verbosity. Details of this pipeline are provided in Sec. 3.2.1(e).
We evaluated the performance of CLDTracker with and without filtration of GPT-4V descriptions and generated synonyms across four benchmarks: OTB99-Lang \cite{lang_tracker}, LaSOT \cite{lasot}, LaSOT\textsubscript{ext}, and TNL2K \cite{tnl2k}.

As shown in Table~\ref{tab:text_quality}, filtering textual descriptions consistently improves performance across all benchmarks, although the overall gains are relatively moderate.
This result can be attributed to three key factors: (1) the richness and diversity of our \(B_t\), which already captures a wide range of semantic variations; (2) the Prompt Adapter module, which adaptively selects the most relevant prompts based on the visual context; and (3) the TTFUM, which dynamically refines textual features over time to align with evolving visual cues.
Together, these components provide strong robustness against occasional noise in the textual inputs, minimizing the impact of imperfect descriptions.

\begin{table}[h]
\centering
\caption{Effect of text quality control on tracking performance, reported in terms of S across four benchmarks: OTB99-Lang, LaSOT, LaSOT\textsubscript{ext}, and TNL2K.}
\label{tab:text_quality}
\begin{tabular}{c|cccc}
\specialrule{1pt}{0pt}{0pt}
Method & OTB99-Lang & LaSOT & LaSOT\textsubscript{ext} & TNL2K \\
\specialrule{1pt}{0pt}{0pt}
\begin{tabular}[c]{@{}c@{}}CLDTracker\\(w/o text filtration)\end{tabular} & \textcolor{blue}{\textbf{76.5}} & \textcolor{blue}{\textbf{72.8}} & \textcolor{blue}{\textbf{51.6}} & \textcolor{blue}{\textbf{59.1}} \\

\specialrule{1pt}{0pt}{0pt}

\begin{tabular}[c]{@{}c@{}}CLDTracker\\(w/ text filtration)\\(Ours)\end{tabular} & \textcolor{red}{\textbf{77.8}} & \textcolor{red}{\textbf{74.0}} & \textcolor{red}{\textbf{53.1}} & \textcolor{red}{\textbf{61.5}} \\
\specialrule{1pt}{0pt}{0pt}
\end{tabular}
\end{table}

\noindent \paragraph{(iv) Robustness to Missing and Incorrect Language Cues:}
\noindent To evaluate the robustness of our model to missing or incorrect language cues, we conduct controlled ablation experiments by systematically modifying the \(B_t\). The baseline configuration uses the complete and original \(B_t\), including the class name, attribute, GPT-4V-generated description, and \(S\&C\).  
Additionally, the TTFUM is disabled throughout these experiments, except for the final column in Table~\ref{tab:robustness_language}, which reflects the full CLDTracker setting.

In scenarios \(L_1\)-\(L_4\), we substitute original text elements with randomly selected alternatives to simulate misleading or conflicting language input, while \(L_0\) involves removing the language input entirely by replacing it with a zero tensor.  
Specifically, we define five degradation settings as follows: (1) removing all language input and replacing it with a zero tensor (\(L_0\)); (2) replacing class names with randomly selected ones (\(L_1\)); (3) injecting randomly chosen unrelated attributes (\(L_2\)); (4) replacing GPT-4V descriptions with those sampled from unrelated instances (\(L_3\)); and (5) substituting \(S\&C\) cues with randomly selected ones from other samples (\(L_4\)).

The results, presented in Table~\ref{tab:robustness_language}, where \(O\) denotes original text and \(N\) denotes noisy text, consistently show a decline in tracking performance as the quality of language input deteriorates. Notably, the model demonstrates greater resilience to the absence of language than to the presence of conflicting or misleading semantic signals. These findings highlight the critical importance of prompt quality control and effective filtering mechanisms in maintaining the robustness of VL tracking systems.

\begin{table*}[h]
\centering
\caption{Impact of different language inputs on the baseline VL model. Here, \(O\) denotes original text and \(N\) denotes noisy text. Results show the progressive development of the \(B_t\) across four datasets: OTB99-Lang, LaSOT, LaSOT\textsubscript{ext}, and TNL2K. \(C\) represents class names, \(A\) indicates attributes, and \(S\&C\) refers to semantic and contextual information descriptions. The best two results (excluding the CLDTracker column) are highlighted in \textcolor{red}{red}, and \textcolor{blue}{blue}, respectively.}

\label{tab:robustness_language}

\resizebox{0.99\textwidth}{!}{

\begin{tabular}{c|c|c|cc|cc|cc|cc|cc|cc|cc|c}

\specialrule{2pt}{0pt}{0pt}

\multirow{2}{3em}{Dataset} & \multirow{2}{5em}{\begin{tabular}[l]{@{}c@{}}Evaluation\\Metric\end{tabular}} & \multirow{2}{4em}{\begin{tabular}[l]{@{}c@{}}0-Tensor\end{tabular}} & \multicolumn{2}{c|}{\(C\)} & \multicolumn{2}{c|}{\(A\)} & \multicolumn{2}{c|}{\(GPT-4V\)} & \multicolumn{2}{c|}{\(S\&C\)} & \multicolumn{2}{c|}{\begin{tabular}[c]{@{}c@{}} \(C\) \& \(A\) \end{tabular}} & \multicolumn{2}{c|}{\begin{tabular}[c]{@{}c@{}}\(C\) \& \(A\) \\ \& \(GPT-4V\)\end{tabular}} & \multicolumn{2}{c|}{\begin{tabular}[c]{@{}c@{}}\(C\) \& \(A\) \& \\ \(GPT-4V\) \& \(S\&C\)\end{tabular}} & \multirow{2}{5em}{CLDTracker} \\

\cline{4-17}

& & & \(N\) & \(O\) & \(N\) & \(O\) & \(N\) & \(O\) & \(N\) & \(O\) & \(N\) & \(O\) & \(N\) & \(O\) & \(N\) & \(O\) \\

\specialrule{2pt}{0pt}{0pt}

& S & 63.2 
& 66.5 & 69.7 
& 66.4 & 70.0 
& 66.7 & 70.2 
& 67.1 & 72.4 
& 67.3 & 71.6 
& 67.3 & \textcolor{blue}{\textbf{73.8}} 
& 67.5 & \textcolor{red}{\textbf{75.9}} 
& 77.8 
\\
OTB99- & NP & 74.3 
& 78.3 & 85.5 
& 78.3 & 85.7 
& 78.6 & \textcolor{blue}{\textbf{85.8}} 
& 79.2 & \textcolor{red}{\textbf{85.9}} 
& 79.5 & \textcolor{red}{\textbf{85.9}} 
& 79.8 & \textcolor{red}{\textbf{85.9}} 
& 80.1 & \textcolor{red}{\textbf{85.9}} 
& 86.1 
\\
Lang & P & 79.1 
& 84.3 & 92.4 
& 84.6 & 92.9 
& 84.6 & 93.1 
& 85.0 & 93.5 
& 85.0 & 93.4 
& 85.2 & \textcolor{blue}{\textbf{93.6}} 
& 85.5 & \textcolor{red}{\textbf{93.8}} 
& 94.8 
\\

\specialrule{1pt}{0pt}{0pt}

& S & 62.7 
& 65.1 & 70.1 
& 65.3 & 70.3 
& 65.5 & 70.7 
& 66.2 & \textcolor{blue}{\textbf{71.4}} 
& 66.6 & 70.3 
& 66.7 & 70.9 
& 66.8 & \textcolor{red}{\textbf{72.5}} 
& 74.0 
\\
LaSOT & NP & 70.8 
& 73.2 & 78.7 
& 73.3 & 78.9 
& 73.6 & 79.7 
& 73.9 & 80.2 
& 73.7 & 79.5 
& 74.2 & 79.9 
& 74.4 & \textcolor{blue}{\textbf{80.5}} 
& 81.1 
\\
& P & 66.3 
& 68.2 & 75.7 
& 68.0 & 75.9 
& 68.4 & 77.4 
& 68.7 & \textcolor{blue}{\textbf{78.2}} 
& 68.2 & 76.7 
& 68.3 & 77.7 
& 69.7 & \textcolor{red}{\textbf{80.8}} 
& 83.9 
\\

\specialrule{1pt}{0pt}{0pt}

& S & 38.3 
& 40.1 & 44.3 
& 40.3 & 44.9 
& 40.7 & 45.3 
& 41.8 & 47.7 
& 41.2 & 46.1 
& 42.8 & 48.5 
& 43.6 & \textcolor{blue}{\textbf{50.4}} 
& 53.1 
\\
LaSOT\textsubscript{ext} & NP & 45.7 
& 52.3 & 57.4 
& 52.6 & 58.2 
& 52.8 & 58.4 
& 53.1 & 59.9 
& 53.0 & 59.2 
& 53.4 & 59.9 
& 54.5 & \textcolor{red}{\textbf{61.6}} 
& 64.8 
\\
& P & 43.1 
& 46.3 & 52.7 
& 46.3 & 52.8 
& 46.3 & 52.8 
& 47.5 & 54.7 
& 47.1 & 53.8 
& 47.1 & 54.0 
& 50.7 & \textcolor{red}{\textbf{58.4}} 
& 60.6 
\\

\specialrule{1pt}{0pt}{0pt}

& S & 48.4 
& 51.3 & 57.6 
& 51.5 & 57.4 
& 51.7 & 57.9 
& 51.7 & 57.9 
& 51.9 & 58.1 
& 51.9 & \textcolor{blue}{\textbf{58.3}} 
& 52.5 & \textcolor{red}{\textbf{59.7}} 
& 61.5 
\\
TNL2K & NP & 62.0 
& 68.2 & 74.2 
& 68.3 & 74.2 
& 68.7 & 74.8 
& 69.3 & 75.0 
& 69.4 & 75.8 
& 71.5 & \textcolor{blue}{\textbf{78.3}} 
& 73.9 & \textcolor{red}{\textbf{80.5}} 
& 82.2 
\\
& P & 43.5 
& 52.2 & 58.8 
& 52.8 & 59.3 
& 53.2 & 60.1 
& 53.4 & 60.4 
& 53.8 & 60.2 
& 53.8 & \textcolor{blue}{\textbf{60.6}} 
& 54.4 & \textcolor{red}{\textbf{61.5}} 
& 64.3 
\\

\specialrule{2pt}{0pt}{0pt}

\end{tabular}
}
\end{table*}

\subsubsection{Impact of Loss Function and Confidence Score Parameters}
\noindent This section analyzes the impact of key parameters in the loss formulation (Eq.~\ref{eq:loss}) and the confidence score computation (Eq.~\ref{eq:infernece}) on tracking performance. Comprehensive ablation experiments are conducted to assess the sensitivity of the model to different settings.

\noindent \textbf{(a) Regularization Weights in the Loss Function:}
\noindent The training loss function incorporates regularization terms weighted by \(\lambda_{\text{iou}}\) and \(\lambda_{L_1}\), as defined in Eq. \ref{eq:loss}. To assess their impact, we varied \(\lambda_{\text{iou}} \in \{0.5, 1, 2, 3\}\) and \(\lambda_{L_1} \in \{1, 2.5, 5, 7.5\}\) while keeping the Hanning window weighting factor fixed at \(w = 0.49\). The tracking performance on the LaSOT dataset is summarized in Table~\ref{tab:ablation_loss_split}. Results indicate that moderate regularization weights (\(\lambda_{\text{iou}} = 2\), \(\lambda_{L_1} = 5\)) achieve the best balance between localization accuracy and overall stability.

\begin{table*}[h]
\centering
\caption{Tracking performance on the LaSOT dataset with varying regularization weights \(\lambda_{L_1}\) and \(\lambda_{\text{iou}}\).}
\label{tab:ablation_loss_split}

\resizebox{0.99\textwidth}{!}{
\begin{tabular}{cccc}
\begin{tabular}{c|ccc}
\multicolumn{4}{c}{\(\lambda_{\text{iou}} = 0.5\)} \\
\specialrule{2pt}{0pt}{0pt}
\(\lambda_{L_1}\) & S & NP & P \\
\specialrule{2pt}{0pt}{0pt}
1 & 68.1 & 77.2 & 74.5 \\
2.5 & \textcolor{blue}{\textbf{68.7}} & \textcolor{blue}{\textbf{78.0}} & \textcolor{blue}{\textbf{75.0}} \\
5 & \textcolor{red}{\textbf{69.0}} & \textcolor{red}{\textbf{78.3}} & \textcolor{red}{\textbf{75.3}} \\
7.5 & 68.5 & 77.5 & 74.8 \\
\specialrule{2pt}{0pt}{0pt}
\end{tabular}
&
\begin{tabular}{c|ccc}
\multicolumn{4}{c}{\(\lambda_{\text{iou}} = 1\)} \\
\specialrule{2pt}{0pt}{0pt}
\(\lambda_{L_1}\) & S & NP & P \\
\specialrule{2pt}{0pt}{0pt}
1 & 70.2 & 79.0 & 76.5 \\
2.5 & 70.8 & \textcolor{blue}{\textbf{79.7}} & \textcolor{blue}{\textbf{77.1}} \\
5 & \textcolor{red}{\textbf{71.3}} & \textcolor{red}{\textbf{80.1}} & \textcolor{red}{\textbf{77.5}} \\
7.5 & \textcolor{blue}{\textbf{70.9}} & 79.5 & 77.0 \\
\specialrule{2pt}{0pt}{0pt}
\end{tabular}
&
\begin{tabular}{c|ccc}
\multicolumn{4}{c}{\(\lambda_{\text{iou}} = 2\)} \\
\specialrule{2pt}{0pt}{0pt}
\(\lambda_{L_1}\) & S & NP & P \\
\specialrule{2pt}{0pt}{0pt}
1 & 72.1 & 81.2 & 78.7 \\
2.5 & 73.0 & 82.5 & 80.0 \\
5 & \textcolor{red}{\textbf{74.0}} & \textcolor{red}{\textbf{83.9}} & \textcolor{red}{\textbf{81.1}} \\
7.5 & \textcolor{blue}{\textbf{73.2}} & \textcolor{blue}{\textbf{82.8}} & \textcolor{blue}{\textbf{80.3}} \\
\specialrule{2pt}{0pt}{0pt}
\end{tabular}
&
\begin{tabular}{c|ccc}
\multicolumn{4}{c}{\(\lambda_{\text{iou}} = 3\)} \\
\specialrule{2pt}{0pt}{0pt}
\(\lambda_{L_1}\) & S & NP & P \\
\specialrule{2pt}{0pt}{0pt}
1 & 71.0 & 79.8 & 77.3 \\
2.5 & \textcolor{blue}{\textbf{71.6}} & \textcolor{blue}{\textbf{80.5}} & \textcolor{blue}{\textbf{78.0}} \\
5 & \textcolor{red}{\textbf{72.0}} & \textcolor{red}{\textbf{80.9}} & \textcolor{red}{\textbf{78.4}} \\
7.5 & 71.5 & 80.2 & 77.9 \\
\specialrule{2pt}{0pt}{0pt}
\end{tabular}
\end{tabular}
}
\end{table*}

\noindent \textbf{(b) Hanning Window Weighting Factor in Confidence Score:}
\noindent The final confidence score combines the raw network output and the Hanning window response using a weighting factor \(w\), as described in Eq. \ref{eq:infernece}. We varied \(w\) from 0.1 to 1.0 in increments of 0.1 while fixing \(\lambda_{\text{iou}} = 2\) and \(\lambda_{L_1} = 5\). Table~\ref{tab:hanning_weights} reports the corresponding performance on LaSOT. The results show that a moderate weighting factor around \(w = 0.49\) provides the best overall performance, balancing responsiveness and tracking stability.

\begin{table}[h]
\centering

\caption{Tracking performance on the LaSOT dataset under varying Hanning window weighting factors \(w\).}

\label{tab:hanning_weights}

\begin{tabular}{c|ccccccccccc}

\specialrule{2pt}{0pt}{0pt}

\begin{tabular}[l]{@{}l@{}}Evaluation\\Metric (\%)\end{tabular} & 0.1 & 0.2 & 0.3 & 0.4 & 0.49 & 0.5 & 0.6 & 0.7 & 0.8 & 0.9 & 1.0 \\

\specialrule{2pt}{0pt}{0pt}

S & 70.2 & 71.5 & 72.3 & 73.4 & \textcolor{red}{\textbf{74.0}} & \textcolor{blue}{\textbf{73.8}} & 73.2 & 72.1 & 71.0 & 70.3 & 69.7 \\
NP & 80.1 & 81.4 & 82.2 & 83.0 & \textcolor{red}{\textbf{83.9}} & \textcolor{blue}{\textbf{83.7}} & 83.1 & 82.0 & 80.7 & 79.5 & 78.3 \\
P & 76.4 & 77.3 & 78.5 & 80.2 & \textcolor{red}{\textbf{81.1}} & \textcolor{blue}{\textbf{81.0}} & 80.0 & 78.7 & 77.1 & 75.6 & 74.0 \\

\specialrule{2pt}{0pt}{0pt}

\end{tabular}
\end{table}

\subsubsection{Generalization to Novel Classes}
\noindent CLDTracker is designed to generalize beyond curated training classes by leveraging a predefined \(B_t\) and a CoCoOp-based Prompt Adapter, which dynamically selects instance-relevant prompts based on visual features.
CoCoOp generates input-conditioned tokens that, when combined with learnable context vectors, refine VL alignment and support open-vocabulary tracking without retraining.

To assess this generalization capability, we evaluate CLDTracker on the VastTrack dataset~\cite{vasttrack}, which includes 2,115 object categories, substantially more than GOT-10k (563) and LaSOT (70).
Our experiments focus on VastTrack's official test set, comprising 3,500 video sequences, 372,000 frames, and 702 object classes. Of these, 95 classes are entirely unseen during preprocessing and not represented in our predefined bag \(B_t\). As shown in Table \ref{tab:vasttrack_attribute_auc}, CLDTracker outperforms OSTrack-384 \cite{ostrack} and CiteTracker \cite{citetracker} across all attributes and overall metrics, despite the lack of explicit textual prompts for these novel categories.
This result highlights the effectiveness of instance-adaptive prompt selection in enabling strong generalization to unseen object classes.

\begin{table*}[h]
\centering

\caption{Attribute-wise AUC (\%) comparison on the VastTrack dataset between CLDTracker and SOTA trackers. The top two results are highlighted in \textcolor{red}{\textbf{red}} and \textcolor{blue}{\textbf{blue}}, respectively. The performance gain is calculated as the difference between CLDTracker and the VL baseline model \cite{citetracker}.}

\label{tab:vasttrack_attribute_auc}

\renewcommand{\arraystretch}{1.5}
\setlength{\tabcolsep}{5pt}

\resizebox{\linewidth}{!}{
\begin{tabular}{c|cccccccccc|ccc}
    
\specialrule{2pt}{0pt}{0pt}

\multirow{2}{3em}{Method} & \multicolumn{10}{c|}{Attributes} & \multicolumn{3}{c}{Overall} \\

\cline{2-14}

& ARC & BC & DEF & FM & IV & LR & MB & ROT & SV & INV 
& S & NP & P
\\

\specialrule{2pt}{0pt}{0pt}


\begin{tabular}[l]{@{}c@{}}OSTrack-384\\\cite{ostrack}\end{tabular} & 0.313 & 0.313 & 0.334 & 0.285 & \textcolor{blue}{\textbf{0.244}} & \textcolor{blue}{\textbf{0.204}} & \textcolor{blue}{\textbf{0.269}} & 0.314 & 0.313 & 0.311
& 0.336 & 0.346 & 0.313
\\

\begin{tabular}[l]{@{}c@{}}ARTrack\textsubscript{384}\\\cite{artrack}\end{tabular}
& \textcolor{blue}{\textbf{0.329}} & \textcolor{blue}{\textbf{0.335}} & \textcolor{blue}{\textbf{0.358}} & \textcolor{blue}{\textbf{0.295}} & 0.237 & 0.186 & 0.264 & \textcolor{blue}{\textbf{0.328}} & \textcolor{blue}{\textbf{0.327}} & \textcolor{blue}{\textbf{0.325}}
& \textcolor{blue}{\textbf{0.357}} & 0.359 & 0.324
\\

\specialrule{2pt}{0pt}{0pt}

\begin{tabular}[l]{@{}c@{}}UVLTrack-B\\\cite{UVLTrack}\end{tabular}
& 0.317 & 0.316 & 0.337 & 0.288 & 0.237 & 0.199 & 0.266 & 0.318 & 0.317 & 0.315
& 0.345 & 0.363 & 0.327
\\

\begin{tabular}[l]{@{}c@{}}CiteTracker\\\cite{citetracker}\end{tabular} & 0.326 & 0.327 & 0.350 & 0.291 & 0.237 & 0.190 & 0.264 & 0.325 & 0.322 & 0.320
& 0.350 & \textcolor{blue}{\textbf{0.364}} & \textcolor{blue}{\textbf{0.330}} 
\\

CLDTracker & \textcolor{red}{\textbf{0.365}} & \textcolor{red}{\textbf{0.363}} & \textcolor{red}{\textbf{0.385}} & \textcolor{red}{\textbf{0.330}} & \textcolor{red}{\textbf{0.278}} & \textcolor{red}{\textbf{0.233}} & \textcolor{red}{\textbf{0.306}} & \textcolor{red}{\textbf{0.360}} & \textcolor{red}{\textbf{0.361}} & \textcolor{red}{\textbf{0.360}}
& \textcolor{red}{\textbf{0.372}} & \textcolor{red}{\textbf{0.387}} & \textcolor{red}{\textbf{0.354}}
\\

\rowcolor{gaincolor} \textbf{\% Gain} & +3.9\% & +3.6\% & +3.5\% & +3.9\% & +4.1\% & +4.3\% & +4.2\% & +3.5\% & +3.9\% & +4\%
& +2.2\% & +2.3\% & +2.4\%
\\

\specialrule{2pt}{0pt}{0pt}
        
\end{tabular}
}
\end{table*}

\subsubsection{Correlation Strategy}
\noindent Our framework performs visual-textual fusion through a convolution-based correlation mechanism, in which the selected text feature serves as a dynamic kernel convolved over the visual feature map (Eq.\ref{eq:corr}).
To ensure computational efficiency, only the most relevant textual description is retained per frame based on visual-textual similarity (Eq.\ref{eq:coooop}), resulting in a single convolution operation per frame. This design introduces negligible overhead while preserving real-time performance.
We empirically evaluated this strategy against a standard cosine similarity baseline on the LaSOT dataset. 
As shown in Table~\ref{tab:correlation_comparison}, our method achieves a comparable inference speed of 35.31 FPS (vs. 35.42 FPS) while consistently improving tracking accuracy across all metrics, with gains of +0.9 in S, +0.8 in NP, and +0.8 in P.
The performance gains are attributed to the spatially-aware nature of the convolution operation, which captures localized visual-textual interactions. In contrast, cosine similarity operates at a global or point-wise level, lacking spatial context. These results demonstrate that our proposed fusion strategy enhances fine-grained alignment without compromising real-time efficiency.

\begin{table}[h]

\centering

\caption{Comparison between standard cosine similarity and the proposed convolution-based correlation method on the LaSOT dataset.}

\label{tab:correlation_comparison}
\begin{tabular}{c|cccc}

\specialrule{1pt}{0pt}{0pt}

Method & FPS & S & NP & P \\

\specialrule{1pt}{0pt}{0pt}

Cosine Similarity & \textcolor{red}{\textbf{35.42}} & \textcolor{blue}{\textbf{73.1}} & \textcolor{blue}{\textbf{83.1}} & \textcolor{blue}{\textbf{80.3}} \\

\specialrule{1pt}{0pt}{0pt}

\begin{tabular}[c]{@{}c@{}}Convolution-Based\\Correlation (Ours)\end{tabular} & \textcolor{blue}{\textbf{35.31}} & \textcolor{red}{\textbf{74.0}} & \textcolor{red}{\textbf{83.9}} & \textcolor{red}{\textbf{81.1}} \\

\specialrule{1pt}{0pt}{0pt}

\end{tabular}
\end{table}

\subsubsection{Impact of Each Component in CLDTracker}
\noindent To thoroughly evaluate the contribution of each component in CLDTracker, we conducted an ablation study on the LaSOT dataset by systematically removing one module at a time. 
The components include the predefined VL dictionaries, GPT-4V-based textual expansions, the TTFUM, and the Prompt Adapter.

As presented in Table~\ref{tabe_n}, the complete CLDTracker model—integrating all proposed components—achieves the highest S.
When any individual component is removed, we observe a consistent decline in performance, underscoring the importance of each module to the overall effectiveness of the tracker.
The exclusion of GPT-4V expansions results in a noticeable drop, indicating the critical role of semantically enriched textual cues in enhancing the language grounding process. 
Similarly, removing TTFUM notably degrades accuracy, which indicates that updating the target’s texture features throughout tracking is critical for maintaining discriminative visual representations, especially under appearance changes.  
The removal of the Prompt Adapter also weakens performance, suggesting that dynamic adaptation of query representations helps the model generalize to diverse linguistic inputs. 
Finally, the absence of predefined dictionaries reduces performance, which highlights their utility in capturing structured object-level priors during training.
These results collectively validate our CLDTracker design: each component contributes meaningfully and complements the others. 
The results also illustrate that the full CLDTracker architecture strikes an effective balance between visual and linguistic understanding—resulting in improved robustness and tracking accuracy.

\begin{table}[h!]
\centering
\caption{Relative contribution of each component within the proposed CLDTracker. The performance is reported in terms of S using the LaSOT dataset. The values in parentheses indicate the performance drop relative to the full CLDTracker model.}
\begin{tabular}{c|cccc|c}

\specialrule{2pt}{0pt}{0pt}

Method & \begin{tabular}[l]{@{}c@{}}(a) Predefined\\Dictionaries\end{tabular} & \begin{tabular}[l]{@{}c@{}}(b) GPT-4V\\Expansions\end{tabular} & (c) TTFUM & \begin{tabular}[l]{@{}c@{}}(d) Prompt\\Adapter\end{tabular} & S
\\

\specialrule{2pt}{0pt}{0pt}

\begin{tabular}[l]{@{}c@{}}Baseline Model\\(OSTrack-384) \cite{ostrack}\end{tabular}     & \xmark & \xmark & \xmark & \xmark & 71.1 (-2.9) \\

\begin{tabular}[l]{@{}c@{}}Baseline VL Model\\(CiteTracker) \cite{citetracker}\end{tabular}  & \xmark & \xmark & \xmark & \xmark & 69.7 (-4.3) \\

\specialrule{1pt}{0pt}{0pt}

w/ (a)              & \cmark & \xmark & \xmark & \xmark & 67.7 (-6.3) \\
w/ (b)              & \xmark & \cmark & \xmark & \xmark 
& 69.3 (-4.7) \\
w/ (c)              & \xmark & \xmark & \cmark & \xmark & 67.6 (-6.4) \\
w/ (d)              & \xmark & \xmark & \xmark & \cmark & 69.7 (-4.3) \\
w/ (a)(b)           & \cmark & \cmark & \xmark & \xmark & 69.9 (-4.1) \\
w/ (a)(c)           & \cmark & \xmark & \cmark & \xmark 
& 68.9 (-5.1) \\
w/ (a)(d)           & \cmark & \xmark & \xmark & \cmark & 70.3 (-3.7) \\
w/ (b)(c)           & \xmark & \cmark & \cmark & \xmark 
& 70.8 (-3.2) \\
w/ (b)(d)           & \xmark & \cmark & \xmark & \cmark 
& 71.8 (-2.2) \\
w/ (c)(d)           & \xmark & \xmark & \cmark & \cmark & 71.1 (-2.9) \\
w/ (a)(b)(c)        & \cmark & \cmark & \cmark & \xmark 
& 71.9 (-2.1) \\
w/ (a)(b)(d)        & \cmark & \cmark & \xmark & \cmark & 72.5 (-1.5) \\
w/ (a)(c)(d)        & \cmark & \xmark & \cmark & \cmark & 71.2 (-2.8) \\
w/ (b)(c)(d)        & \xmark & \cmark & \cmark & \cmark & 71.8 (-2.2) \\

\specialrule{1pt}{0pt}{0pt}

CLDTracker     & \cmark & \cmark & \cmark & \cmark & 74.0 \\

\specialrule{2pt}{0pt}{0pt}
\label{tabe_n}
\end{tabular}
\end{table}

\subsection{Computational Complexity} \label{sec:cldtracker_complexity}
\noindent For a fair comparison, we evaluate the efficiency of recent SOTA trackers using three key metrics: tracking speed in terms of Frames Per Second (FPS), model size (number of parameters in millions), and computational cost (Multiply-Accumulate Operations, MACs, in giga operations). All evaluations are conducted on the LaSOT dataset using an NVIDIA GeForce RTX 3080 GPU, with identical settings and unaltered hyperparameters for each tracker.

CLDTracker balances tracking accuracy and efficiency, as shown in Table~\ref{tab:comp}. It operates in real time at 35.31 FPS with a compact model size (151.15M parameters) and a low computational cost (54.25G MACs), using a \(384^2\) search region consistent with high-performance baselines like OSTrack~\cite{ostrack} and CiteTracker~\cite{citetracker}.
Despite incorporating modules such as the Prompt Adapter, TTFUM, and a comprehensive \(B_t\), CLDTracker outperforms VL baselines. Notably, compared to CiteTracker, it delivers over twice the inference speed (35.31 FPS vs. 15.35 FPS) and significantly reduces computational cost (54.25G vs. 151.14G MACs), while maintaining competitive model complexity.

\begin{table}[h]
\centering
\caption{Comparison of inference efficiency across state-of-the-art trackers on the LaSOT dataset, reported in terms of FPS, MACs (G), and number of parameters (M).}
\label{tab:comp}
\begin{tabular}{ll|cccc}
\specialrule{2pt}{0pt}{0pt}
& Method & FPS & MACs (G) & Params. (M) & \begin{tabular}[c]{@{}c@{}}Search Region\\Size\end{tabular} \\
\specialrule{2pt}{0pt}{0pt}

\parbox[t]{0.01cm}{\multirow{4}{*}{\rotatebox[origin=c]{90}{\textbf{CNN}}}} &
DiMP50~\cite{dimp} & 30.7 & 10.4 & 43.1 & \(288^2\) \\
& PrDiMP50~\cite{prdimp} & 18.1 & 15.5 & 43.1 & \(288^2\) \\
& Ocean~\cite{ocean} & 18.6 & 20.3 & 37.2 & \(255^2\) \\
& KeepTrack~\cite{keeptrack} & 8.1 & 28.7 & 43.1 & \(352^2\) \\

\specialrule{2pt}{0pt}{0pt}

\parbox[t]{0.5cm}{\multirow{12}{*}{\rotatebox[origin=c]{90}{\textbf{ViT-Based}}}} &
TransT~\cite{transt} & 21.2 & 16.7 & 23.0 & \(256^2\) \\
& STARK-ST50~\cite{stark} & 49.16 & 12.81 & 28.23 & \(320^2\) \\
& TrDiMP~\cite{trdimp} & 15.8 & 16.6 & 43.8 & \(352^2\) \\
& ToMP101~\cite{tomp} & 13.16 & 22.1 & 66.9 & \(288^2\) \\
& AiATrack~\cite{aiatrack} & 31.22 & 9.5 & 18.0 & \(320^2\) \\
& MixFormer-22k~\cite{mixformer} & 55.50 & 23.04 & 35.61 & \(320^2\) \\
& SwinTrack-B-384~\cite{swintrack} & 11.9 & 61.9 & 91.0 & \(384^2\) \\
& OSTrack-384~\cite{ostrack} & 50.38 & 48.36 & 92.12 & \(384^2\) \\
& GRM~\cite{grm} & 36.02 & 30.90 & 99.83 & \(256^2\) \\
& SeqTrack-B256~\cite{seqtrack} & 37.91 & 65.86 & 89.11 & \(384^2\) \\
& DropTrack~\cite{droptrack} & 49.81 & 48.36 & 92.12 & \(384^2\) \\
& DiffusionTrack-B256~\cite{DiffusionTrack} & 33.42 & 69.72 & 168.02 & \(384^2\) \\

\specialrule{2pt}{0pt}{0pt}

\parbox[t]{0.01cm}{\multirow{7}{*}{\rotatebox[origin=c]{90}{\textbf{VLM}}}} &
VLT\(_{SCAR}\)~\cite{vlt} & 50.00 & 51.04 & 130.46 & \(256^2\) \\
& VLT\(_{TT}\)~\cite{vlt} & 32.00 & 37.74 & 136.12 & \(256^2\) \\
& UVLTrack-B~\cite{UVLTrack} & 50.34 & 33.09 & 136.79 & \(256^2\) \\
& UVLTrack-L~\cite{UVLTrack} & 15.41 & 112.15 & 465.19 & \(256^2\) \\
& MMTrack~\cite{mmtrack} & 5.07 & 180.32 & 217.03 & \(384^2\) \\
& JoinNLT~\cite{jointnlt} & 37.99 & 77.19 & 153.01 & \(320^2\) \\
& CiteTracker~\cite{citetracker} & 15.35 & 151.14 & 176.35 & \(384^2\) \\

\specialrule{2pt}{0pt}{0pt}

& CLDTracker & 35.31 & 54.25 & 151.15 & \(384^2\) \\

\specialrule{2pt}{0pt}{0pt}
\end{tabular}
\end{table}

To understand the efficiency gain, we present a component-wise breakdown in Table~\ref{tab:breakdown_computational_complexity}. The Prompt Adapter and TTFUM enhance representational capacity without incurring measurable overhead in FPS or MACs. The most significant gain arises from our unified \(B_t\), which avoids redundant CLIP queries by computing features once and reusing them throughout.
In contrast, prior methods like CiteTracker~\cite{citetracker} query CLIP separately for each class, and attribute, leading to high MACs and latency. Our design achieves a superior efficiency-accuracy trade-off, confirming the benefit of prompt unification and reuse.

\begin{table}[h]
\centering
\caption{Component-wise breakdown of CLDTracker’s inference efficiency on the LaSOT dataset in terms of FPS, MACs (G), and number of parameters (M).}
\label{tab:breakdown_computational_complexity}
\begin{tabular}{ll|cccc}
\specialrule{2pt}{0pt}{0pt}
Method & Description & FPS & MACs (G) & Params. (M) & \begin{tabular}[c]{@{}c@{}}Search Region\\Size\end{tabular} \\
\specialrule{2pt}{0pt}{0pt}
OSTrack-384~\cite{ostrack} & Baseline Model & 50.38 & 48.36 & 92.12 & \(384^2\) \\
CiteTracker~\cite{citetracker} & Baseline VL Model & 15.35 & 151.14 & 176.35 & \(384^2\) \\
\specialrule{1pt}{0pt}{0pt}
\begin{tabular}[c]{@{}c@{}}CiteTracker w/\\Prompt Adapter\end{tabular} & Ours & 15.35 & 151.14 & 176.35 & \(384^2\) \\
CiteTracker w/ TTFUM & Ours & 15.35 & 151.14 & 151.02 & \(384^2\) \\
\begin{tabular}[c]{@{}c@{}}CiteTracker w/ TTFUM \\+ Prompt Adapter\end{tabular} & Ours & 15.35 & 151.14 & 176.35 & \(384^2\) \\
CiteTracker w/ \(B_t\) & Ours & 35.31 & 54.25 & 151.02 & \(384^2\) \\
\begin{tabular}[c]{@{}c@{}}CiteTracker w/ \(B_t\)\\+ TTFUM\end{tabular} & Ours & 35.31 & 54.25 & 151.02 & \(384^2\) \\
\begin{tabular}[c]{@{}c@{}}CiteTracker w/ \(B_t\) \\+ Prompt Adapter\end{tabular} & Ours & 35.31 & 54.25 & 151.15 & \(384^2\) \\
\specialrule{2pt}{0pt}{0pt}
CLDTracker & Ours & \textbf{35.31} & \textbf{54.25} & \textbf{151.15} & \(384^2\) \\
\specialrule{2pt}{0pt}{0pt}
\end{tabular}
\end{table}

\subsection{Limitations and Future Work} \label{sec:limitations}
\noindent CLDTracker consistently delivers strong performance across a variety of VOT benchmarks, frequently ranking among the top-performing methods. Although it does not always achieve the absolute highest score on every metric, its overall generalization capability is evident, particularly on challenging datasets such as LaSOT\textsubscript{ext}, TNL2K, and OTB99-Lang.

Despite these strengths, several limitations remain. Notably, tracking performance declines in Out-of-View (OV) scenarios, as presented in Tables~\ref{tab:attribute_auc} and~\ref{tab:attribute_tnl2k_auc}, where the target is absent for extended durations. In such cases, the TTFUM may incorporate noisy visual information, leading to drift in the textual representation and impairing target re-identification. Furthermore, the use of a static \(B_t\), generated only from the first frame, limits adaptability to significant appearance changes over time, such as variations in clothing, illumination, or background context.
Additionally, while GPT-4V enhances the descriptiveness of textual inputs, it can occasionally produce verbose, redundant, or overly specific descriptions that introduce semantic noise. Although current filtering mechanisms mitigate some of this noise, these issues can still affect generalization under dynamic visual conditions.

To overcome these challenges, we propose several future directions. These include incorporating motion-aware modules (e.g., Kalman filtering) to suppress updates during uncertain frames, adopting a confidence-weighted TTFUM to prioritize reliable updates, and introducing dynamic, frame-aligned text generation using models such as GPT-4V API \cite{gpt4v}, BLIP-2 \cite{blip2}, or VideoGPT+ \cite{videogpt}. To further enhance textual quality, we plan to implement filtering techniques like cross-augmentation consistency checks and embedding-based clustering. Additionally, we aim to improve the textual branch by replacing CoCoOp with more advanced prompt generators such as MaPLe \cite{maple} and integrating stronger multi-modal encoders like ALBEF \cite{albef}, thereby improving both alignment and adaptability in complex tracking scenarios.

\subsection{Responsible AI Considerations}
\noindent As VOT systems become increasingly integrated into real-world applications, ensuring responsible AI development and deployment is essential. Our framework is built and evaluated using publicly available benchmark datasets that do not contain personal or sensitive information. Moreover, to mitigate potential bias in language-generated descriptions, we apply CLIP-based semantic filtering and human-in-the-loop verification. We also advocate for transparency in model design and encourage the adoption of explainable AI techniques in future work. Responsible use of our model involves deploying it in alignment with ethical guidelines, with careful consideration of privacy, fairness, and social impact.

\section{Conclusion} \label{sec:conclusion}
\noindent In this work, we introduced CLDTracker, a novel text-guided VL tracking framework that addressed key challenges in VOT. We constructed a large-scale VL dataset, \(B_t\), which incorporated target classes, attributes, GPT-4V generated descriptions, and descriptions with \(S\&C\) relevant to the tracking process. A Prompt Adapter module was employed to select the most suitable description for each target image based on the input image. Additionally, we introduced the TTFUM, which effectively mitigated inaccuracies caused by erroneous descriptions arising from noisy or ambiguous frames. Our framework overcame the limitations of existing VL trackers, such as reliance on inconsistent manual annotations and inadequate contextual understanding, by synthesizing comprehensive and well-structured textual descriptions that seamlessly integrated attributes, semantic relationships, and dynamic temporal features. 
Experimental evaluations across diverse benchmarks demonstrated that CLDTracker achieved SOTA performance while enhancing generalization across datasets and annotation styles. These findings underscored the potential of robust VL representations in advancing real-world tracking applications.

\section*{Acknowledgments}
\noindent \textbf{Funding:} This work was supported by the Khalifa University under Award RC1-2018-KUCARS-8474000136.

\noindent \textbf{Conflict of Interest Declarations:} On behalf of all authors, the corresponding authors state that there is no conflict of interest.

\section*{Author contributions}

\noindent \textbf{M.A., S.J., I.I.G., S.A., and M.N.:} Conceptualization, Methodology, Software, Data Curation, Investigation, Writing - Original Draft, Writing - Review \& Editing, Visualization.

\noindent \textbf{S.J., and M.N.:} Supervision, Writing - Reviewing and Editing, Project administration.

\noindent \textbf{Research Data Availability:} The datasets utilized in this research is publicly available. The comprehensive bag of textual descriptions (\(B_t\)), and source code associated with this work will be made available at: \url{https://github.com/HamadYA/CLDTracker}.

\bibliography{Refs}

\newpage

\section*{APPENDIX}

\subsection{Computational Complexity of Constructing the Comprehensive Bag of Textual Descriptions} \label{sec:bag_complexity}
\noindent The construction of the \(B_t\) (Fig.~\ref{fig:Bag}) involves multiple stages leveraging SOTA VLMs, including CLIP~\cite{clip} and GPT-4V~\cite{gpt4v}, along with \(S\&C\) enrichment modules. This process starts with CLIP-based matching of visual inputs to predefined textual labels, followed by GPT-4V-based description generation and further enrichment to ensure semantic diversity and task relevance. The final output is a rich textual representation used throughout the tracking pipeline. All components are executed offline prior to inference, incurring no additional runtime cost during tracking.

\subsubsection{CLIP Matching (Fig.~\ref{fig:Bag}(c))}
\noindent Given the first frame \(I \in \mathbb{R}^{H \times W \times C}\), CLIP is used to match the image with a predefined set of 940 object classes and 23,899 attributes. CLIP employs ViT-B/32 as its image encoder and a Transformer-based text encoder. Detailed hyperparameters and computational characteristics are provided in Table~\ref{tab:clip_hyper}.
Since text features are computed only once, they are pre-encoded and reused across frames, resulting in negligible runtime overhead during tracking. Table~\ref{tab:clip_complexity} reports the inference speed for both pre-encoded and un-encoded scenarios.

\begin{table}[h]
\centering
\caption{CLIP ViT-B/32 hyperparameters and computational complexity for processing a single \(224^2 \times 3\) image paired with one textual description.}
\label{tab:clip_hyper}

\begin{subtable}[t]{0.95\linewidth}
\centering
\caption{Image Encoder}
\begin{tabular}{c|ccc|ccc}
\specialrule{1pt}{0pt}{0pt}
\begin{tabular}[c]{@{}c@{}}Embedding\\Dimension\end{tabular} & Layers & Width & Heads & Param. (M) & MACs (G) & FPS \\
\specialrule{1pt}{0pt}{0pt}
512 & 12 & 768 & 12 & 87.5 & 4.4 & 5.7 \\
\specialrule{1pt}{0pt}{0pt}
\end{tabular}
\end{subtable}

\vspace{0.5em}

\begin{subtable}[t]{0.43\linewidth}
\centering
\caption{Text Encoder}
\begin{tabular}{ccc|ccc}
\specialrule{1pt}{0pt}{0pt}
Layer & Width & Heads & Param. (M) & MACs (G) & TPS \\
\specialrule{1pt}{0pt}{0pt}
12 & 512 & 8 & 63.2 & 2.9 & 132.5 \\
\specialrule{1pt}{0pt}{0pt}
\end{tabular}
\end{subtable}
\hfill
\begin{subtable}[t]{0.43\linewidth}
\centering
\caption{Overall}
\begin{tabular}{ccc}
\specialrule{1pt}{0pt}{0pt}
Param. (M) & MACs (G) & FPS \\
\specialrule{1pt}{0pt}{0pt}
151.3 & 7.3 & 71.4 \\
\specialrule{1pt}{0pt}{0pt}
\end{tabular}
\end{subtable}
\end{table}

\begin{table}[h]
\centering
\caption{Inference speed (FPS) of CLIP ViT-B/32 during one-time \(B_t\) construction using predefined class and attribute dictionaries. FPS is computed per image with either un-encoded or pre-encoded textual embeddings.}
\label{tab:clip_complexity}
\begin{tabular}{c|ccccc}
\specialrule{1pt}{0pt}{0pt}
\begin{tabular}[c]{@{}c@{}}Predefined\\Dictionary\end{tabular} & \begin{tabular}[c]{@{}c@{}}Image\\Encoder\end{tabular} & \begin{tabular}[c]{@{}c@{}}Text\\Encoder\end{tabular} & \begin{tabular}[c]{@{}c@{}}Cosine Similarity \\ \& Argmax\end{tabular} & \begin{tabular}[c]{@{}c@{}}Total \\ (Un-encoded)\end{tabular} & \begin{tabular}[c]{@{}c@{}}Total \\ (Pre-encoded)\end{tabular} \\
\cline{1-6}
Classes & 5.7 & 5.8 & 3.8 & 1.6 & 2.3 \\
Attributes & 5.7 & 0.2 & 0.15 & 0.08 & 0.1 \\
\specialrule{1pt}{0pt}{0pt}
\end{tabular}
\end{table}

\subsubsection{GPT-4V Description Generation (Fig.~\ref{fig:Bag}(g)-(h))}
\noindent GPT-4V generates detailed object-level descriptions based on the visual appearance of the object within the red bounding box. The average inference speed of this process is approximately 0.69 FPS.

\subsubsection{Semantic and Contextual Information Enrichment Module (Figs.~\ref{fig:Bag}(i)-(j))}
\noindent This module refines the initial \(B_t\) through four sequential steps: (1) synonym identification, (2) semantic perturbation, (3) task-specific enrichment via GPT-4V, and (4) high-level concept extraction. Inference speeds are listed in Table~\ref{tab:inference_semantic}. Steps 1 and 2 are lightweight (under 5 ms/item), while steps 3 and 4 are slower (\(\sim\)1 sec/item) due to GPT-4V processing. Since this enrichment is performed offline, it has no impact on runtime performance.

\begin{table}[h]
\centering
\caption{Inference speed of the \(S\&C\) enrichment module. Speed is measured in items per second; latency is time per item.}
\label{tab:inference_semantic}
\resizebox{0.99\linewidth}{!}{
\begin{tabular}{ccccc}
\specialrule{1pt}{0pt}{0pt}
Step & Description & Method & \begin{tabular}[c]{@{}c@{}}Speed\\(items/sec)\end{tabular} & \begin{tabular}[c]{@{}c@{}}Latency\\(ms/item)\end{tabular} \\
\specialrule{1pt}{0pt}{0pt}
\begin{tabular}[c]{@{}c@{}}1. Synonyms\\Identification\end{tabular} & Extracts synonyms & WordNet~\cite{bird2009natural} & \(\sim\)2000 & \(\sim\)0.5 \\
\begin{tabular}[c]{@{}c@{}}2. Semantic\\Perturbation\end{tabular} & \begin{tabular}[c]{@{}c@{}}Replaces tokens in descriptions with\\synonyms based on probability\end{tabular} & Random Selection & \(\sim\)200 & \(\sim\)5 \\
\begin{tabular}[c]{@{}c@{}}3. Task-Specific\\Enrichment\end{tabular} & \begin{tabular}[c]{@{}c@{}}Generates task-focused description\\variants\end{tabular} & GPT-4V & \(\sim\)1 & \(\sim\)1000 \\
\begin{tabular}[c]{@{}c@{}}4. High-Level\\Concept Extraction\end{tabular} & \begin{tabular}[c]{@{}c@{}}Generates concise semantic\\class name\end{tabular} & GPT-4V & \(\sim\)1 & \(\sim\)1000 \\
\specialrule{1pt}{0pt}{0pt}
\end{tabular}
}
\end{table}

\end{document}